\documentclass{article}

\usepackage{microtype}
\usepackage{enumitem}
\usepackage{graphicx}
\usepackage{booktabs} 

\usepackage{hyperref}

\usepackage[accepted]{icml2025}

\usepackage{amsmath}
\usepackage{amssymb}
\usepackage{mathtools}
\usepackage{amsthm}
\usepackage{graphicx}

\usepackage{amsmath}
\usepackage{amssymb}
\usepackage{mathtools}
\usepackage{amsthm}
\usepackage{tabularx}       
\usepackage{makecell}
\usepackage{amsmath,amsfonts}
\usepackage{bm}
\usepackage{caption}
\usepackage{subcaption}
\usepackage{graphicx}
\usepackage{enumitem}
\usepackage{caption}

\usepackage{listings}
\usepackage[capitalize,noabbrev]{cleveref}

\usepackage{url}

\theoremstyle{plain}

\theoremstyle{definition}

\theoremstyle{remark}

\icmltitlerunning{Revisiting End-To-End Sparse Autoencoder Training: A Short Finetune Is All You Need}

\begin{document}

\twocolumn[
\icmltitle{Revisiting End-To-End Sparse Autoencoder Training: A Short Finetune Is All You Need}


\begin{icmlauthorlist}
\icmlauthor{Adam Karvonen}{aff1}
\end{icmlauthorlist}

\icmlaffiliation{aff1}{Independent}

\icmlcorrespondingauthor{Adam Karvonen}{adam.karvonen@gmail.com}

\icmlkeywords{Sparse Autoencoders, Machine Learning, ICML}

\vskip 0.3in
]

\printAffiliationsAndNotice{} 

\begin{abstract}
    Sparse autoencoders (SAEs) are widely used for interpreting language model activations. A key evaluation metric is the increase in cross-entropy loss between the original model logits and the reconstructed model logits when replacing model activations with SAE reconstructions. Typically, SAEs are trained solely on mean squared error (MSE) when reconstructing precomputed, shuffled activations. Recent work introduced training SAEs directly with a combination of KL divergence and MSE (“end-to-end” SAEs), significantly improving reconstruction accuracy at the cost of substantially increased computation, which has limited their widespread adoption. We propose a brief KL+MSE fine-tuning step applied only to the final 25M training tokens (just a few percent of typical training budgets) that achieves comparable improvements, reducing the cross-entropy loss gap by 20–50\%, while incurring minimal additional computational cost. We further find that multiple fine-tuning methods (KL fine-tuning, LoRA adapters, linear adapters) yield similar, non-additive cross-entropy improvements, suggesting a common, easily correctable error source in MSE-trained SAEs. We demonstrate a straightforward method for effectively transferring hyperparameters and sparsity penalties between training phases despite scale differences between KL and MSE losses. While both ReLU and TopK SAEs see significant cross-entropy loss improvements, evaluations on supervised SAEBench metrics yield mixed results, with improvements on some metrics and decreases on others, depending on both the SAE architecture and downstream task. Nonetheless, our method may offer meaningful improvements in interpretability applications such as circuit analysis with minor additional cost.
\end{abstract}

\section{Introduction}

\begin{figure*}
    \centering
    \includegraphics[width=0.85\linewidth]{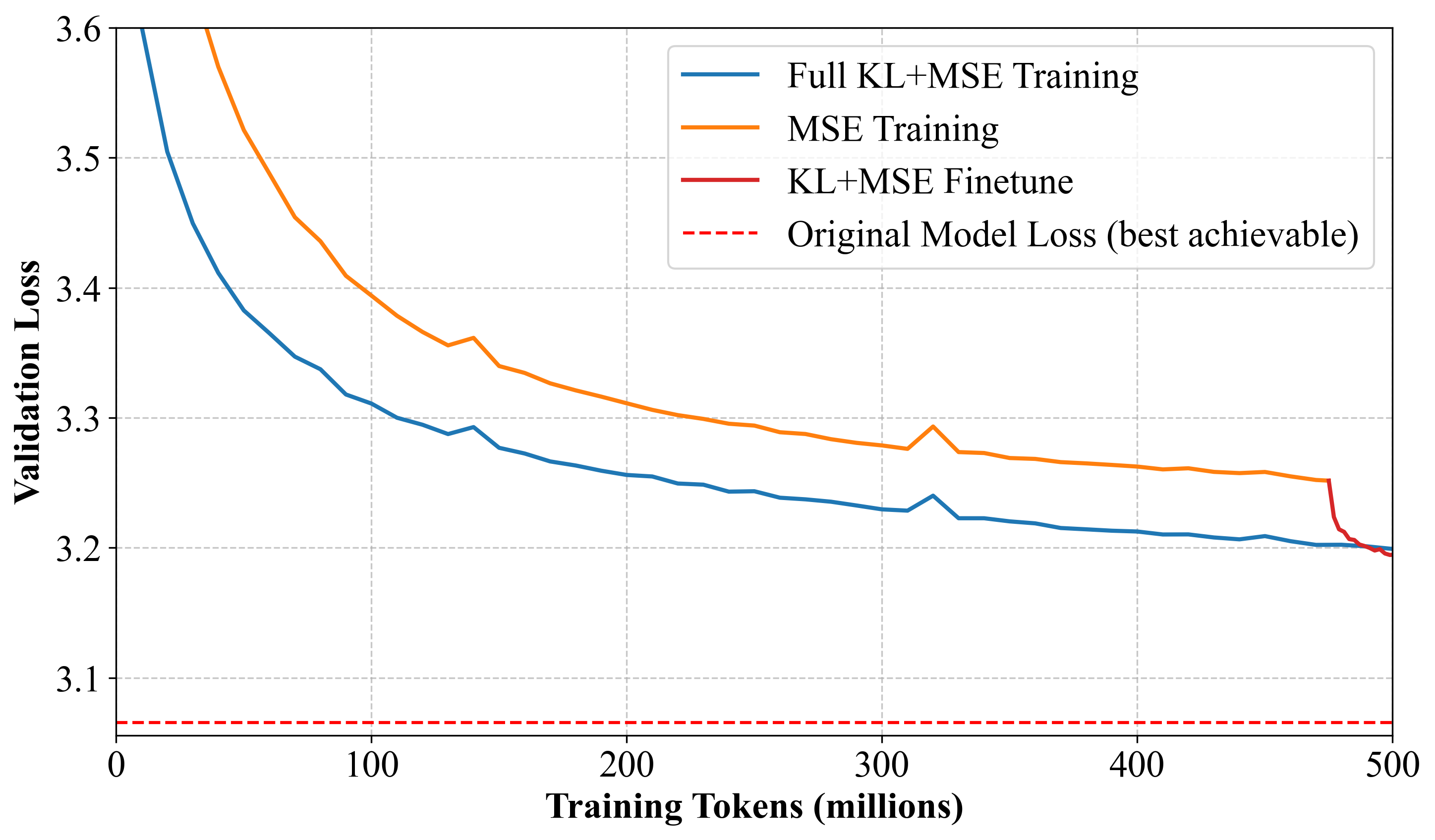}
    \caption{
    Comparison of training approaches for a sparse autoencoder (K=80, width=16K) on Pythia-160M. The proposed KL+MSE fine-tuning approach (25M tokens) achieves slightly better performance than full end-to-end (E2E) training \cite{braun2024identifyingfunctionallyimportantfeatures} on the same amount of data while reducing wall-clock time by approximately 50\%.
    }

    \label{fig:figure_1}
\end{figure*}

Sparse autoencoders (SAEs) have emerged as an important tool in mechanistic interpretability, enabling the decomposition of language model activations into sparse linear combinations of interpretable latent features \cite{cunningham2023sparseautoencodershighlyinterpretable, bricken2023monosemanticity}. The central hypothesis behind SAEs is that neural activations can be effectively represented using sparse combinations of meaningful latent directions, facilitating deeper understanding and interpretability of neural network computations.

The primary evaluation metric for SAEs is the increase in cross-entropy loss incurred when original model activations are replaced by SAE reconstructions during inference, capturing the trade-off between sparsity and fidelity. Recent advances have focused extensively on improving this sparsity-fidelity trade-off through novel SAE architectures \cite{rajamanoharan2024improving, mudide2024efficient}, activation functions \cite{gao2024scaling, taggart2024prolu, rajamanoharan2024jumping, bussmann2024batchtopk, ayonrinde2024adaptivesparseallocationmutual}, and training loss formulations \cite{karvonen2024measuringprogressdictionarylearning, bussmann2024matryoshka}.

Typically, SAEs are trained with mean squared error (MSE) loss when reconstructing precomputed and shuffled model activations \cite{lieberum2024gemmascopeopensparse}. However, this approach does not directly optimize the cross-entropy loss between the original model logits and reconstructed model logits used during evaluation. Recent methods have explored training SAEs with a combination of KL divergence and MSE loss to align training objectives more closely with evaluation objectives. Notably, \citet{braun2024identifyingfunctionallyimportantfeatures} proposed "end-to-end" SAE training, using KL+MSE loss throughout training. While effective, this method substantially increases computational cost, thus limiting practical applicability and widespread adoption by the field.

Alternatively, \citet{chen2025lowrankadaptingmodelssparse} introduced a method leveraging low-rank adapters (LoRA) to fine-tune the underlying language model around pre-trained SAEs. This approach achieves similar performance gains with much lower computational overhead. However, it introduces additional complexity and alters the original language model, which may be undesirable for interpretability studies.

In this work, we revisit end-to-end SAE training and demonstrate a simpler yet equally effective strategy: applying a brief fine-tuning stage with KL+MSE loss for only the final 25M tokens of training (0.5-10\% of typical training budgets (250M - 5B tokens)). In our particular setting, this targeted approach exceeds the performance of both full end-to-end training and LoRA adapters, while significantly reducing wall-clock time—approximately 50\% reduction compared to full end-to-end training, and a minor reduction compared to LoRA adapters. In other common scenarios, such as training on larger language models, employing early stopping during activation collection, or amortizing SAE training across precomputed activations, wall-clock time savings relative to end-to-end training can easily exceed 90\%. \footnote{For example, training six SAEs at the middle layer involves KL divergence loss (two forward passes + one backward, 4× cost), no early stopping (2×), and no amortization across SAEs (6×), totaling up to 48× extra compute for activation collection. Actual wall-clock savings depend on SAE-to-LLM size ratios.}

These results suggest that training with an MSE loss function learns meaningful features with an easily correctable KL divergence error source which can be removed by multiple methods, of which SAE fine-tuning is the simplest and most performant. We present a practical recipe for smoothly transferring hyperparameters and sparsity penalties from MSE-only to KL+MSE training phases, addressing challenges arising from significant scale differences between these losses.

We evaluate our proposed method on SAEBench \cite{karvonen2025saebenchcomprehensivebenchmarksparse}, a comprehensive suite of metrics beyond reconstruction accuracy. Evaluations on downstream supervised metrics yield mixed results, showing both improvements and declines depending on both the specific evaluation metric and SAE architecture, with particularly pronounced changes for ReLU-based SAEs. Despite this, we believe that achieving considerable reductions in cross-entropy loss at minimal cost may substantially benefit interpretability-focused applications, such as circuit analysis, by reducing the influence of reconstruction error nodes and thus decreasing the risk of missing critical signals.

Our main contributions include:

\begin{enumerate}
\item A simplified and computationally efficient fine-tuning approach for achieving end-to-end SAE training benefits without altering model architecture or adding significant complexity.
\item A practical method for transferring hyperparameters and sparsity penalties between MSE and KL+MSE training phases.
\item Empirical evidence demonstrating mixed results in supervised interpretability metrics on SAEBench, highlighting that interpretability benefits from KL+MSE training depend on both SAE architecture and the downstream task.
\end{enumerate}

We release code, data, and models at \url{github.com/adamkarvonen/sae_kl_finetune}.

\section{Related Work}

\subsection{Sparse Autoencoders for Interpretability}
\label{sec:sparse_autoencoders_for_interpretability}

Sparse autoencoders (SAEs) have gained popularity as an unsupervised interpretability method for analyzing activations of large language models. An SAE generally consists of an encoder-decoder structure: the encoder transforms the original activations into a higher-dimensional but sparse latent representation, while the decoder reconstructs the original activations from this sparse representation. The first widely used architecture involved a linear encoder layer, a sparsity-inducing activation (often ReLU), and a linear decoder. Training these autoencoders involved minimizing a reconstruction loss along with an $L_1$ sparsity regularization applied to the latent representation. Formally, the SAE forward pass and optimization objective can be defined as:

\begin{align}
h &= \text{ReLU}(W_E x + b_E) \\
\hat{x} &= W_D h + b_D \\
\mathcal{L} &= \underbrace{\|x - \hat{x}\|_2^2}_{\text{reconstruction}} + \lambda \underbrace{\|h\|_1}_{\text{sparsity}}
\end{align}

where $x$ is the input activation vector, $h$ represents the sparse latent representation, and $\hat{x}$ is the reconstructed activation vector. The parameters $W_E, b_E$ and $W_D, b_D$ correspond to the weights and biases of the encoder and decoder, respectively, and $\lambda$ is the sparsity penalty which controls the balance between reconstruction accuracy and sparsity.

Since reconstructions by SAEs are inherently imperfect at any given sparsity level, substituting these approximations back into the original model generally results in increased loss. Consequently, recent research has focused extensively on improving SAE architectures and activation functions to enhance reconstruction accuracy at fixed sparsity levels. Prominent advancements include TopK and BatchTopK activation functions \cite{gao2024scaling,bussmann2024batchtopk}, which explicitly control sparsity levels, and JumpReLU \cite{rajamanoharan2024jumping}, which utilizes straight-through estimators to directly optimize an average $L_0$ sparsity objective. These innovations have significantly enhanced the reconstruction accuracy achievable at specific sparsity levels, effectively shifting and improving the sparsity-fidelity frontier. Most recently proposed SAE variants primarily focus on improving the sparsity-fidelity trade-off.

\paragraph{End to End SAE Training}

The recent introduction of "end-to-end" SAE training by \citet{braun2024identifyingfunctionallyimportantfeatures} leverages KL+MSE loss throughout training, directly aligning training with evaluation objectives, and significantly improves the sparsity-fidelity trade-off for a fixed dataset size. However, this approach substantially increases wall-clock time and computational cost—typically requiring two forward passes per optimization step (original and modified models) and additional backward passes, quadrupling total compute relative to conventional SAE training. Furthermore, end-to-end training introduces several operational constraints: it prevents activation shuffling (which has been shown to improve performance \citet{lieberum2024gemmascopeopensparse}), eliminates the ability to amortize activation collection costs across multiple SAE training runs, and prevents the use of early stopping at intermediate layers during activation collection. Depending on the specific architecture and training setup, these limitations can result in an order of magnitude increase in total computational requirements.

Alternatively, \citet{chen2025lowrankadaptingmodelssparse} proposed a method to train on KL divergence with a fixed, pre-trained SAE, through low-rank adaptation (LoRA), efficiently fine-tuning underlying model parameters to reduce reconstruction-induced performance gaps. While computationally cheaper, this strategy introduces additional complexity and modifies the original language model weights, which some may view as undesirable for interpretability studies.

\section{Methods}

Many sparse autoencoder (SAE) training methods, including ReLU-based SAEs, incorporate various auxiliary losses or sparsity penalties. In contrast, TopK-based SAEs enforce strict sparsity deterministically and thus rely solely on mean squared error (MSE) for training. To effectively transfer experimentally determined hyperparameters during the fine-tuning stage, it is crucial to balance these existing losses with any newly introduced loss terms. Therefore, we include both ReLU-based SAEs, with auxiliary sparsity penalties, and TopK-based SAEs, without auxiliary penalties, in our experiments.

While a simpler ReLU-based SAE formulation was described in \cref{sec:sparse_autoencoders_for_interpretability}, our experiments use an improved sparsity penalty following \citet{anthropic_sae_2024}. Specifically, the sparsity penalty for each feature is weighted by the L1 norm of its corresponding decoder vector:

\begin{align}
\mathcal{L} &= \underbrace{|x - \hat{x}|_2^2}_{\text{reconstruction}} + \lambda \underbrace{\sum_i h_i \|w_i\|_1}_{\text{weighted sparsity}}
\end{align}

where $w_i$ is the $i$-th column of the decoder matrix $W_D$.

\subsection{TopK-based Sparse Autoencoders}

TopK-based SAEs differ from ReLU-based SAEs primarily through their sparsity enforcement mechanism. While ReLU SAEs encourage sparsity through an explicit L1 penalty, TopK SAEs enforce a strict sparsity constraint by allowing exactly the top $K$ encoder activations to remain active per input. Specifically, the TopK SAE forward pass and loss are defined as follows:

\begin{align}
h &= \text{TopK}(W_E x + b_E, K) \\
\hat{x} &= W_D h + b_D \\
\mathcal{L} &= |x - \hat{x}|_2^2
\end{align}

where the TopK operation retains only the largest $K$ values in the encoder's output, setting all other values to zero. Thus, sparsity is deterministic, removing the need for an explicit sparsity penalty. However, other auxillary loss penalties are still commonly used, such as the auxk loss for preventing dead features from \citet{gao2024scaling}.

\subsection{Training Method}

After initial training with MSE loss on shuffled activations, we propose performing a brief fine-tuning using a combined KL divergence and MSE loss. This approach follows insights from prior end-to-end SAE work \citet{braun2024identifyingfunctionallyimportantfeatures}, which found that jointly optimizing KL and MSE losses is important for achieving optimal performance. Specifically, training solely on KL divergence tends to degrade MSE reconstruction quality because multiple plausible activation paths exist through the model. Conversely, using only MSE loss ignores the model's actual predictive priorities as measured by KL divergence. By training on both KL divergence and MSE simultaneously, SAEs can capture both accurate reconstruction and meaningful features without substantial degradation in either objective. We also found that using a mixture of KL and MSE loss is more effective than using only KL loss (experimental results in Appendix \ref{app:kl_mse_vs_kl}).

Our fine-tuning occurs for only the final 25 million training tokens, representing merely 0.5-10\% of typical SAE training budgets (250M - 5B tokens). We train on both Gemma-2-2B \citet{gemmateam2024gemma2improvingopen} and Pythia-160M \citet{biderman2023pythiasuiteanalyzinglarge}. The combined loss explicitly aligns SAE training with the evaluation objective—minimizing the increase in cross-entropy loss when original model activations are replaced by SAE reconstructions:

\begin{align}
\mathcal{L}_{\text{finetune}} &= (|x - \hat{x}|_2^2 + \alpha \cdot \text{KL}(f(\hat{x}), f(x))) * 0.5
\end{align}

Here, $f(\cdot)$ denotes the original language model's output probabilities given activations $x$. To ensure the smooth transfer of experimentally determined hyperparameters, we dynamically adjust the KL divergence term scaling factor, $\alpha_{KL}$, at each batch:

\begin{align}
\alpha_{KL} = \frac{\text{MSE loss}}{\text{KL loss} + 1e^{-8}}
\end{align}

This adjustment balances the ratio of KL vs MSE and ensures that the total reconstruction loss (KL divergence combined with MSE) maintains the scale of the original MSE loss. This approach contrasts with the approach of \citet{braun2024identifyingfunctionallyimportantfeatures}, which manually tuned a fixed $\beta$ hyperparameter to balance these losses. Our dynamic adjustment eliminates the need for this additional hyperparameter—an important consideration given that the ratio between MSE and KL losses can vary substantially (we observed from 50-1000×) depending on the model, layer, and whether activations are normalized.

Although $\alpha_{KL}$ exhibits minor fluctuations from batch to batch, we prioritized maintaining a consistent ratio between reconstruction loss and any auxiliary or sparsity losses. We believe exact batch-level balancing is preferable to smoothing or averaging $\alpha_{KL}$ across multiple batches. In Appendix \ref{app:kl_mse_vs_kl}, we observed no degradation to the final loss when applying this dynamic balancing.

We found that learning rate decay is important in achieving optimal performance during fine-tuning. We started with a relatively low learning rate of 5e-5 and applied linear decay to 0 over the fine-tuning period. This short fine-tuning strategy maintains the computational advantages of traditional MSE-based training while slightly exceeding the performance gains of full end-to-end training.

\subsection{Implementation}

In practice, the dynamic balancing of losses described above is implemented succinctly in PyTorch as follows:

\begin{lstlisting}[language=Python]
    alpha_kl = (mse_loss / (kl_loss + 1e-8)).detach()
    loss = (mse_loss + alpha_kl * kl_loss) * 0.5
\end{lstlisting}

\section{Results}

\begin{figure*}
    \centering
    \includegraphics[width=0.85\linewidth]{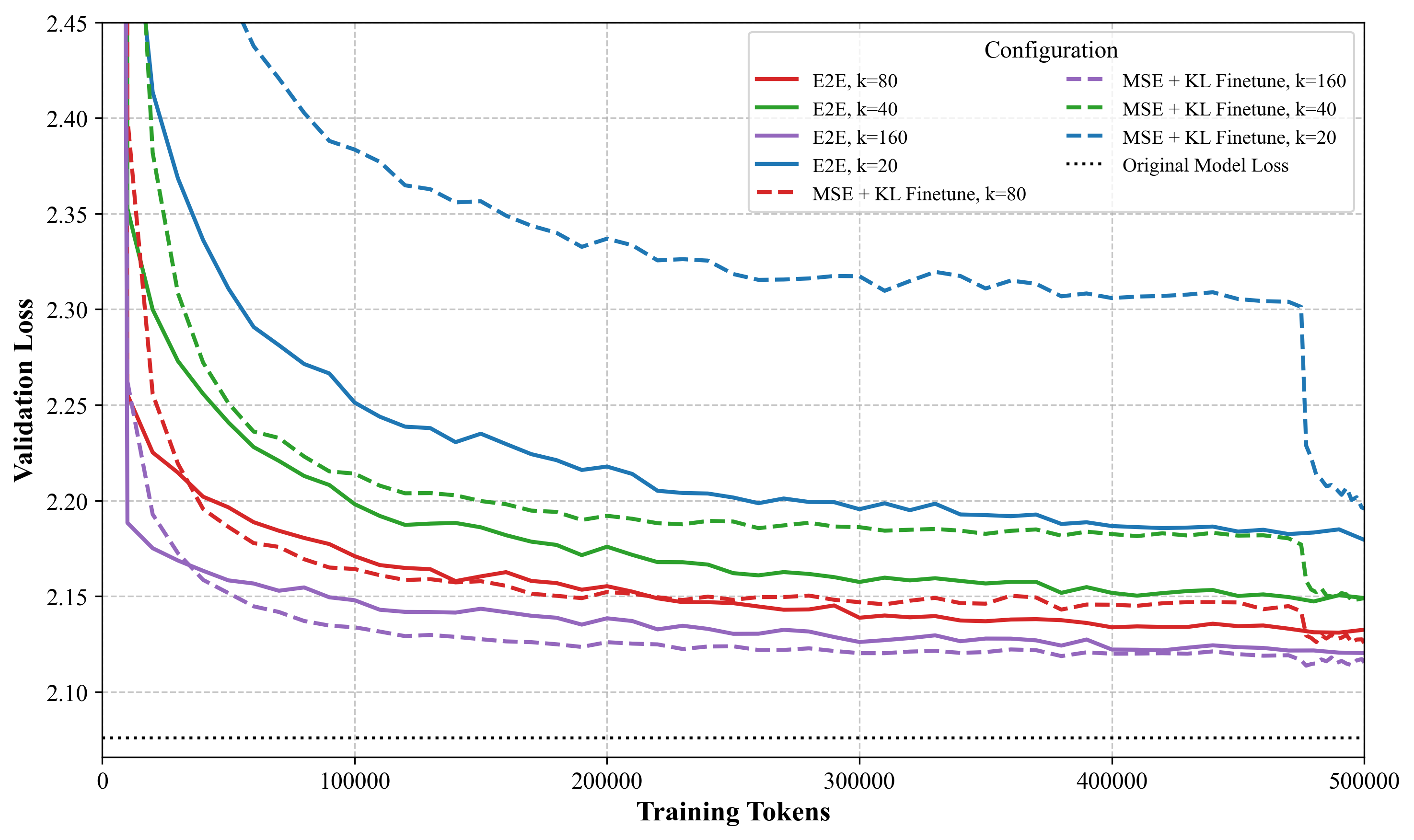}
    \caption{
    Comparison of KL+MSE finetuning (25M tokens) vs full end-to-end training (E2E) on Gemma-2-2B with 65K width SAEs.
    }
    \label{fig:kl_finetune_vs_e2e}
\end{figure*}

\subsection{Cross Entropy Loss Results}

\textbf{Training SAEs on MSE only identifies features capable of achieving competitive cross-entropy loss reductions through various fine-tuning methods; these methods do not stack, making brief KL+MSE fine-tuning optimal for both performance and simplicity.}

We evaluated three different adaptation techniques—LoRA adapting, full KL+MSE fine-tuning, and linear adapters with skip connections—to assess their effectiveness in reducing KL divergence. Individually, each method successfully reduced cross-entropy loss. However, we observed minimal incremental improvements when combining these methods, suggesting the presence of a common, easily correctable source of error in SAEs initially trained only on MSE. Further evidence for this hypothesis comes from our finding that even an extremely low-rank LoRA adapter (rank 2) applied directly to the SAE captures more than half of the benefit achievable through full fine-tuning (Appendix~\ref{app:sae_lora_experiment}). This would also explain why a short KL+MSE fine-tuning stage is sufficient to achieve the benefits of full end-to-end training.

Consequently, our results indicate that training primarily on MSE and then briefly fine-tuning with KL+MSE is the most practical and efficient strategy for optimizing cross-entropy loss. However, we do see differences on supervised SAEBench metrics, which we discuss in the next section.

Below, we detail empirical results supporting these observations.

\textbf{KL+MSE finetuning typically meets or exceeds the performance of full end-to-end training while significantly reducing wall-clock time}. Figure \ref{fig:kl_finetune_vs_e2e} compares our KL+MSE fine-tuning approach against full end-to-end (E2E) training across different sparsity levels (K=20, 40, 80, and 160) on Gemma-2-2B. For most sparsity levels (K=40, 80, 160), our fine-tuning approach matches or exceeds the performance of full E2E training while reducing wall-clock time by approximately 50\%. The actual reduction in wall-clock time will vary depending on several factors, including the ratio of SAE size to LLM size, whether early stopping is employed during activation collection, and whether activation collection costs are being amortized across multiple SAEs.

However, for the highly sparse K=20 configuration, the fine-tuning approach performs worse and shows incomplete convergence, indicating that either additional training data or learning rate tuning may be necessary when operating at high sparsity levels. Interestingly, at K=160, MSE fine-tuning alone achieves comparable or slightly better results than KL+MSE E2E training. We speculate that this could be due to the fact that MSE only is a simpler optimization objective which is sufficient at low sparsities. This may also be due to the fact that these experiments use identical data orderings for fair comparison between methods, meaning they do not benefit from shuffled activations, which could impact training dynamics. However, we still notice a benefit to the KL+MSE finetune for K=160.

\begin{figure*}
    \centering
    \includegraphics[width=0.85\linewidth]{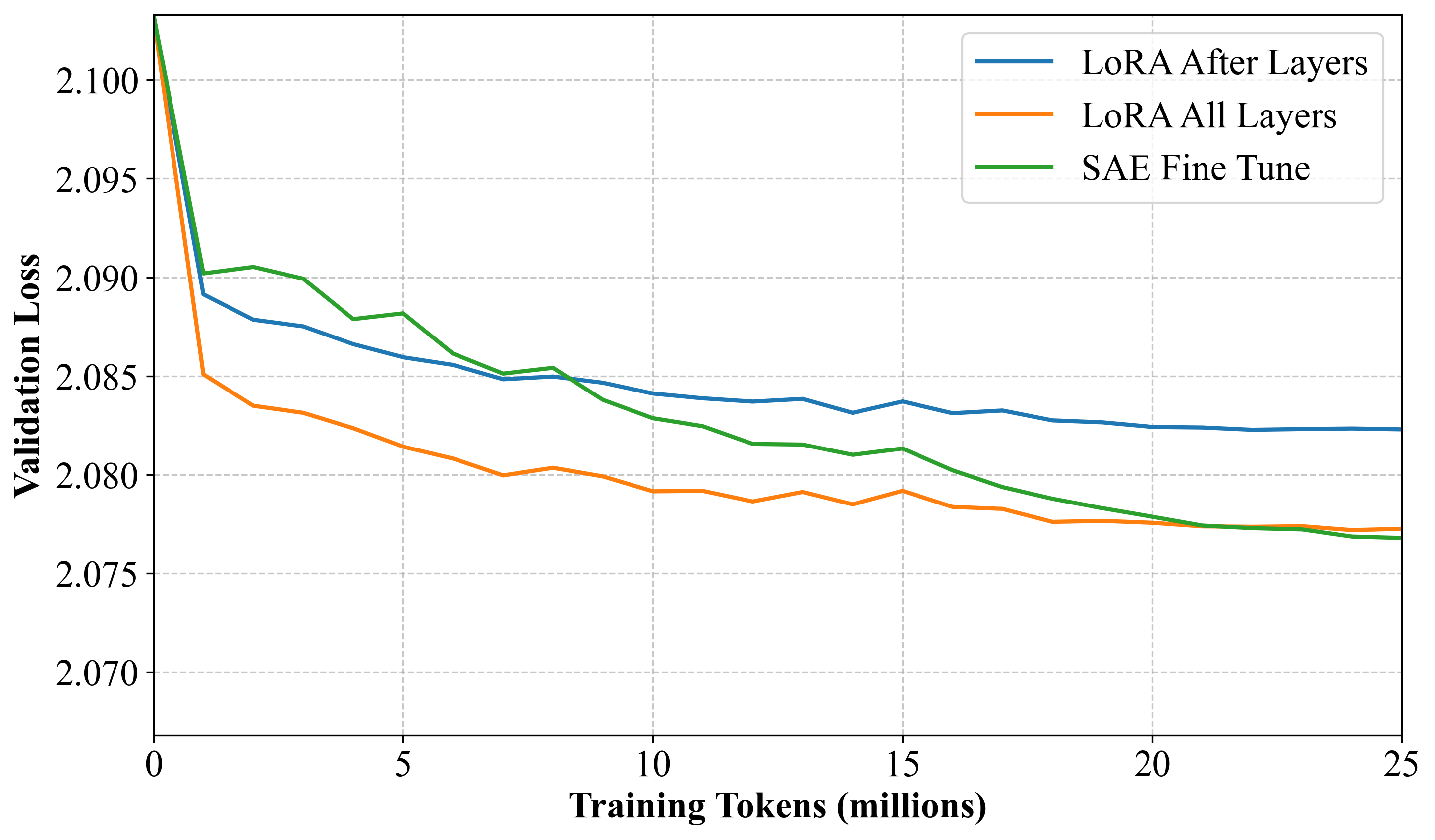}
    \caption{
    Comparison of KL+MSE finetuning (25M tokens) vs LoRA adapters on Gemma-2-2B with 65K width SAEs.
    }
    \label{fig:lora_sae_comparison}
\end{figure*}

\textbf{KL+MSE fine-tuning exceeds the performance of training LoRA adapters on the LLM}. Prior work has proposed two primary methods involving LoRA adapters \cite{chen2025lowrankadaptingmodelssparse}: training adapters across all layers of the language model, and training adapters solely after the SAE layer. Training LoRA adapters across all model layers restricts applicability mainly to TopK-based SAEs, as adapters placed before non-TopK SAEs can "cheat" by decreasing sparsity due to the non-differentiability of L0 sparsity objectives. Under these conditions, our proposed SAE fine-tuning method achieves slightly better performance than training full-model LoRA adapters.

Alternatively, applying LoRA adapters after the SAE avoids the sparsity concerns but leads to significantly weaker performance compared to our SAE fine-tuning approach. Additionally, there are interpretability considerations: many interpretability workflows prefer analyzing the original, unmodified base model, making SAE fine-tuning clearly preferable.

\textbf{Alternative lightweight adaptations can individually capture significant portions of SAE fine-tuning performance, but their benefits fail to stack.} We further investigated alternative lightweight adaptations post-SAE, such as adding low-rank linear transforms or small MLP layers with skip connections after the encoder. Although these methods individually achieved meaningful reductions in cross-entropy loss—approximately 60\% of the gains seen from SAE fine-tuning—we found that combining these adaptations sequentially after fine-tuning provided minimal incremental improvement. This suggests that these methods primarily capture overlapping benefits, making simple SAE fine-tuning preferable for practical applications due to reduced complexity. Full details and experiments are provided in Appendix~\ref{app:alternative_adapter_experiments}.

\begin{figure*}
    \centering
    \includegraphics[width=0.85\linewidth]{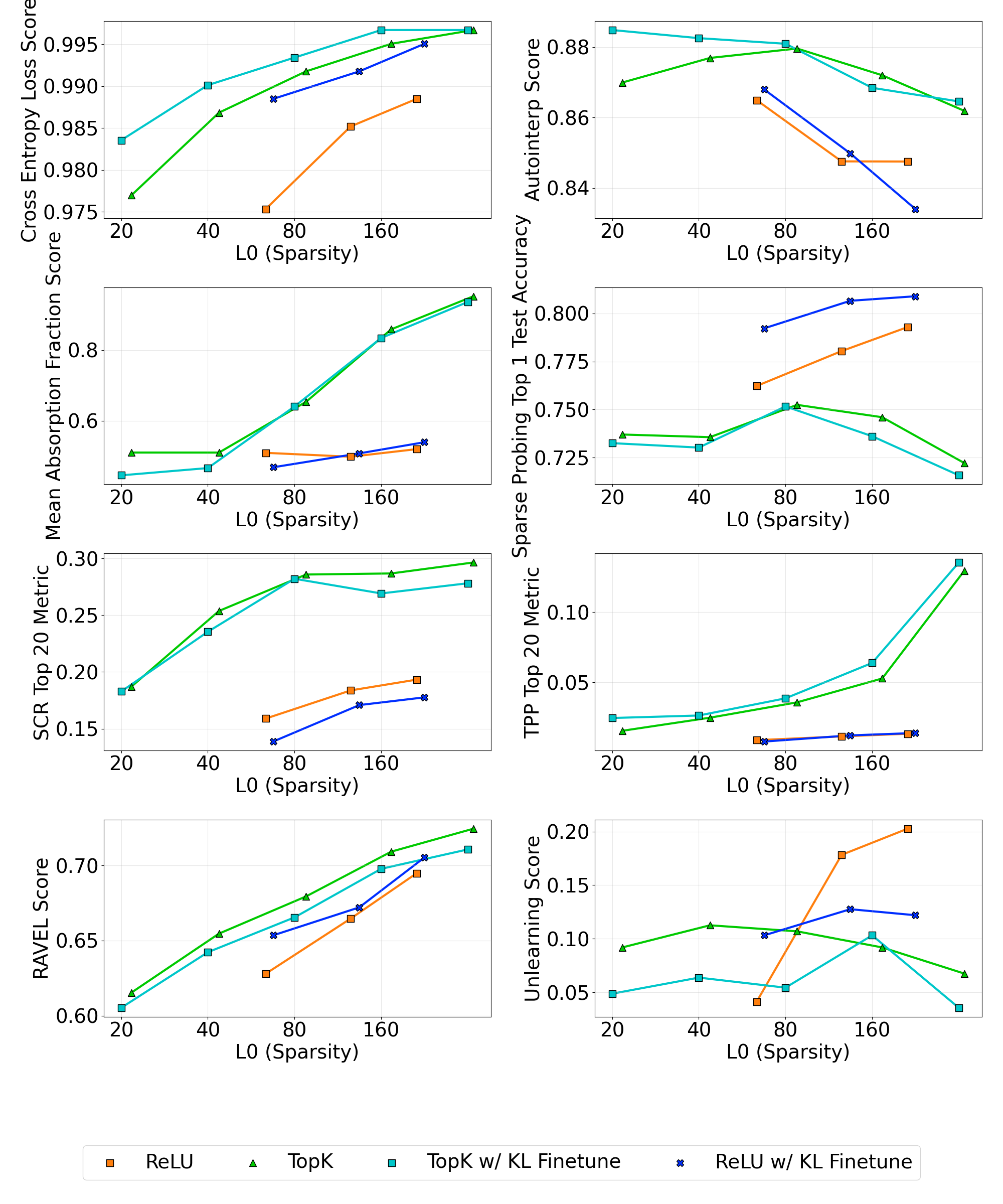}
    \caption{
    SAEBench results for KL+MSE finetuning (15M tokens) on 65k width SAEs on Gemma-2-2B.
    }
    \label{fig:saebench_65k}
\end{figure*}

\subsection{SAEBench Results}

\textbf{Fine-tuning yields mixed results on SAEBench metrics, dependent on both SAE architecture and evaluation metric.} While enhancing the sparsity-fidelity trade-off and reducing cross-entropy loss are important objectives, they ultimately serve as proxy metrics for evaluating SAE utility in practical interpretability tasks. Recently, evaluation suites like SAEBench have emerged to assess SAE performance across diverse downstream interpretability tasks. Several tasks are of note:

\emph{Spurious Correlation Removal (SCR)} measures the ability to debias a spurious correlation from a classifier by ablating identified latents. \emph{Targeted Probe Perturbation (TPP)} measures the ability to degrade the performance of a targeted probe by ablating identified latents, with higher scores indicating that the SAE has latents which correspond to identified concepts. \emph{Sparse Probing} assesses whether individual SAE latents correlate strongly with identified concepts by training probes on single latents, where higher scores indicate better concept quality. \emph{Automated Interpretability} uses an LLM judge to quantify the human-interpretability of selected latents using the \emph{Detection score} proposed by \citet{paulo2024automaticallyinterpretingmillionsfeatures}. \emph{RAVEL} (Resolving Attribute-Value Entanglements in Language Models) measures how cleanly SAEs disentangle related attributes by testing whether targeted interventions on latents can selectively modify model outputs with respect to specific attributes (e.g., changing a city's location) while preserving other related knowledge (e.g., the language spoken there).

We evaluate our fine-tuning method using SAEBench and find substantial improvements in supervised metrics, notably sparse probing and RAVEL scores, for ReLU-based SAEs, as illustrated in Figure \ref{fig:saebench_65k} and Appendix \ref{app:saebench_results}. However, we observe notable decreases in other metrics, such as SCR, suggesting potential trade-offs between different interpretability objectives. In contrast, fine-tuned TopK-based SAEs show smaller and more inconsistent changes across SAEBench metrics at both 16K and 65K widths—improving performance on SCR and TPP metrics, while decreasing on RAVEL and unlearning metrics.

These experiments involved fine-tuning SAEs from the SAEBench baseline suite for 15 million tokens. While our method largely maintains sparsity levels during fine-tuning of ReLU-based SAEs through hyperparameter transfer, we found that a simple dynamic adjustment of the sparsity penalty (detailed in Appendix \ref{app:relu_sparsity_penalty_adjustments}) helps ensure consistent L0 sparsity across experiments.

\textbf{E2E SAEs achieve similar cross-entropy scores but differ significantly on SAEBench metrics.} Interestingly, as shown in Figure \ref{fig:saebench_from_scratch}, E2E-trained SAEs exhibit notably lower performance across several SAEBench metrics, including RAVEL, Feature Absorption, TPP, and SCR, despite comparable cross-entropy scores. This discrepancy suggests that E2E SAEs might identify fundamentally different features capable of achieving similar KL divergence. Additionally, we observe modest performance degradation in TPP and SCR scores during KL fine-tuning.

We hypothesize this degradation may be due to correlations between in-batch datapoints in the absence of activation shuffling, an effect that could potentially be mitigated by using much larger batch sizes. Future work could systematically investigate this by conducting a batch size sweep for E2E SAE training. For now, our primary analysis focuses on fine-tuning the suite of SAEBench SAEs, which are trained on shuffled activations, reflecting current best practices.

\section{Discussion}

\textbf{Why do we observe significant differences between ReLU and TopK SAEs on SAEBench metrics?} We propose two hypotheses that could explain this observation. First, ReLU-based SAEs generally exhibit significantly worse reconstruction accuracy, as reflected by higher cross-entropy loss, compared to TopK-based SAEs. This reduced reconstruction accuracy might disproportionately impact downstream interpretability evaluations.

Second, ReLU-based SAEs are known to suffer from the phenomenon known as "shrinkage", which is the systematic underestimation of feature activations caused by the L1 penalty pushing activations towards zero. \citep{wright2024addressing}. Shrinkage may cause ReLU SAEs to learn qualitatively different—and potentially less faithful—features compared to those identified by TopK SAEs. As a result, ReLU SAEs may have greater room for improvement when fine-tuned using KL divergence, which directly aligns reconstruction with the model's predictive priorities.

\textbf{For applications such as circuit analysis, improved reconstruction accuracy can directly enhance interpretability.} For example, \citet{marks2024sparsefeaturecircuitsdiscovering} employed attribution patching to quantify the contribution of each SAE feature to a metric of interest. Imperfect reconstruction necessitates incorporating an additional “error node” into the analysis, which is often among the most important contributors to the metric’s outcome. This introduces uncertainty about what crucial information might remain hidden within the error node.

A recent effort  \citep{ameisen2025circuit} explores an even more ambitious direction: constructing “replacement models” that substitute interpretable components for entire MLP layers, using cross-layer transcoders (a variant of SAEs). While promising, these models currently match the original model’s output in only around 50\% of cases, even for relatively simple behaviors. While we do not conduct circuit-level evaluations in this work, we believe the significant reduction in KL divergence achieved by our method is likely to improve the fidelity of such replacement models or attribution-based analyses.

\section{Limitations}

\textbf{Mixed results on SAEBench indicate that the practical benefits of KL+MSE fine-tuning vary depending on SAE architecture and evaluation metrics.} Our findings demonstrate that while some interpretability metrics improve, others decline, highlighting potential trade-offs and suggesting the need for careful consideration of architecture-specific and task-specific contexts when employing KL-based fine-tuning.

\textbf{Interpretability metrics may not fully capture the internal representations that models actually use.} Current automated interpretability metrics, including those provided by SAEBench, focus primarily on practical downstream interpretability tasks and tend to emphasize disentangled and easily interpretable features. However, these metrics may not fully capture the internal representations that models actually use, which could be inherently more entangled or complex. For example, it is possible that end-to-end trained SAEs, despite lower scores on SAEBench, might learn representations that are more faithful to the model's internal computations but inherently less interpretable by current automated metrics.

\section{Conclusion}

In this work, we demonstrated that a short KL+MSE fine-tuning stage applied at the end of sparse autoencoder (SAE) training substantially improves cross-entropy loss, achieving results comparable to full end-to-end training but at significantly reduced computational cost. This fine-tuning approach could be particularly beneficial in applications where reconstructive accuracy directly enhances interpretability, such as circuit analysis. Despite these clear improvements in reconstruction fidelity, evaluations on SAEBench metrics produced mixed results, indicating potential trade-offs between different interpretability objectives. Consequently, practitioners adopting KL+MSE fine-tuning should carefully consider their specific interpretability goals and downstream tasks. Future research could further clarify the relationship between SAE training objectives and interpretability outcomes, guiding the development of more effective and interpretable feature representations.

\section*{Acknowledgements}
Adam Karvonen is supported by a grant from Open Philanthropy and a compute grant from Lambda Labs. The author would also like to thank Neel Nanda, Can Rager, Samuel Marks, Bart Bussmann, Patrick Leask, Josh Engels, and Matthew Chen for helpful discussions and feedback.


\clearpage
\bibliographystyle{icml2025}
\bibliography{references}

\newpage
\appendix
\onecolumn

\clearpage
\section{Experiment Details}
\label{app:experiment_details}

\subsection{Training from Scratch vs. End-to-End}

For comparisons between training from scratch (MSE + KL fine-tuning) and full end-to-end (E2E) training, all SAEs were trained on a total of 500 million tokens. In the fine-tuning approach, KL+MSE training was performed only on the final 5\% (25 million tokens) of the training budget. We employed a constant learning rate of $5\times 10^{-5}$. All SAEs were trained with the same dataset, using identical data ordering, a sequence length of 1024, and a batch size of 1. To facilitate stable and fair comparisons without retuning the learning rate, we dynamically balanced the magnitude of the KL and MSE losses, ensuring that the total magnitude of the KL+MSE loss closely matched that of the original MSE-only loss.

We conducted experiments across two model scales. For Gemma-2-2B, we trained SAEs with width 65K at four different sparsity levels (K=20, 40, 80, 160). For Pythia-160M, we trained a single configuration with K=80 and width=16K to validate our findings on a smaller model. We did not apply early stopping to the MSE-only training phase, which could have halved the cost of activation collection.

\subsection{Fine-Tuning SAEBench SAEs}

For fine-tuning the SAEBench SAEs, we trained for 15 million tokens using a sequence length of 1024 and a batch size of 2. We used an initial learning rate of $5\times10^{-5}$, linearly decayed to 0 throughout training.

\subsection{LoRA Adapter Comparison}

When comparing our SAE fine-tuning approach against LoRA adapters, training was performed for 25 million tokens with a sequence length of 1024 and a batch size of 2. We used an initial learning rate of $5\times10^{-5}$, linearly decayed to 0 by the end of training. Our LoRA adapters were rank 16.

All additional comparisons (LoRA adapters, linear adapters, KL-only training) were conducted using the SAEBench baseline TopK SAE with K=80 and width 65K, which was trained on layer 12 of the Gemma-2-2B model. Our training dataset was the Pile, matching the SAEBench training dataset.

\section{KL+MSE vs. KL only}
\label{app:kl_mse_vs_kl}

\begin{figure*}[!htb]
    \centering
    \begin{subfigure}[b]{0.48\textwidth}
        \centering
        \includegraphics[width=\textwidth]{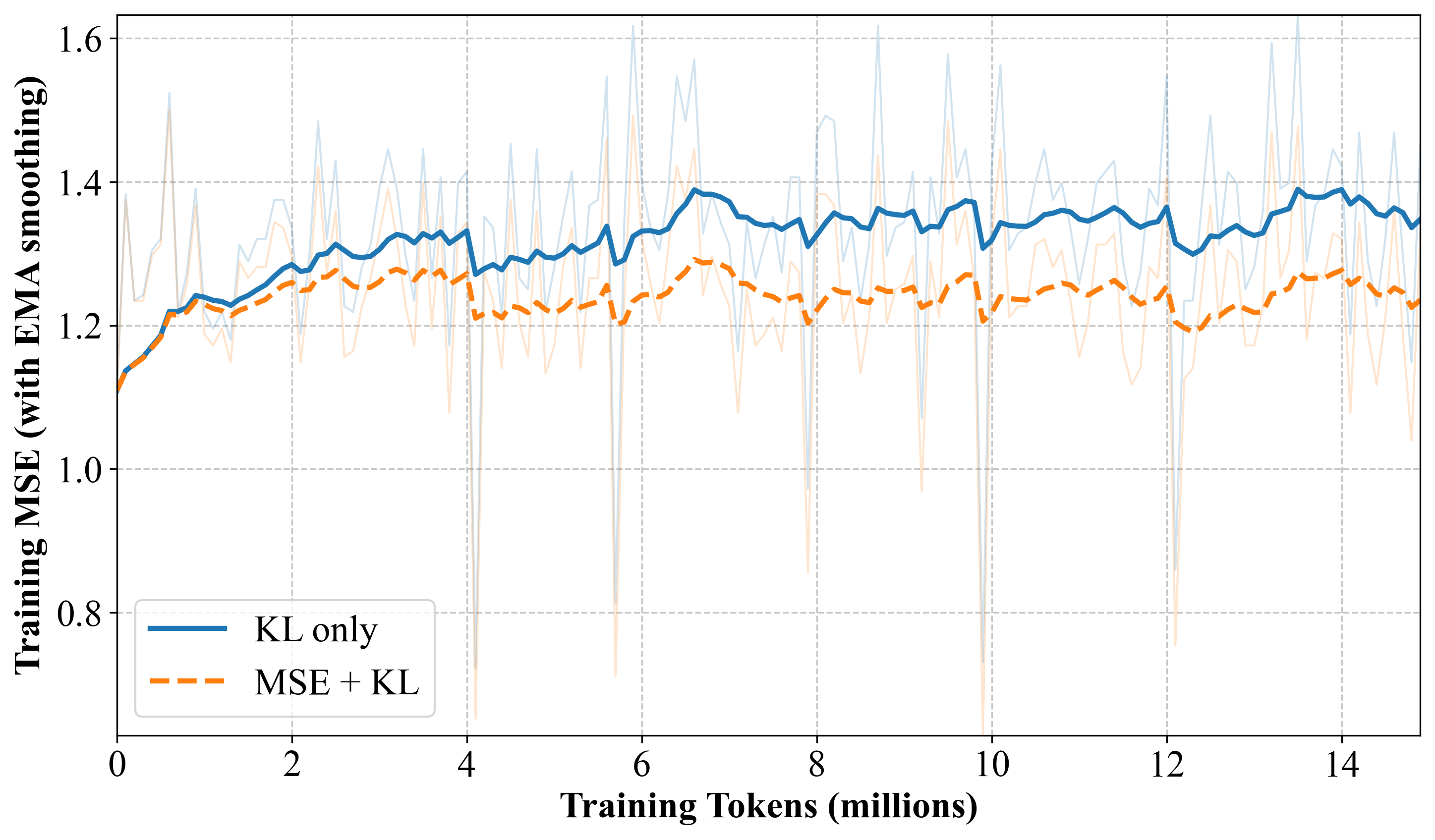}
        \caption{Training MSE comparison}
        \label{fig:kl_mse_vs_kl_a}
    \end{subfigure}
    \hfill
    \begin{subfigure}[b]{0.48\textwidth}
        \centering
        \includegraphics[width=\textwidth]{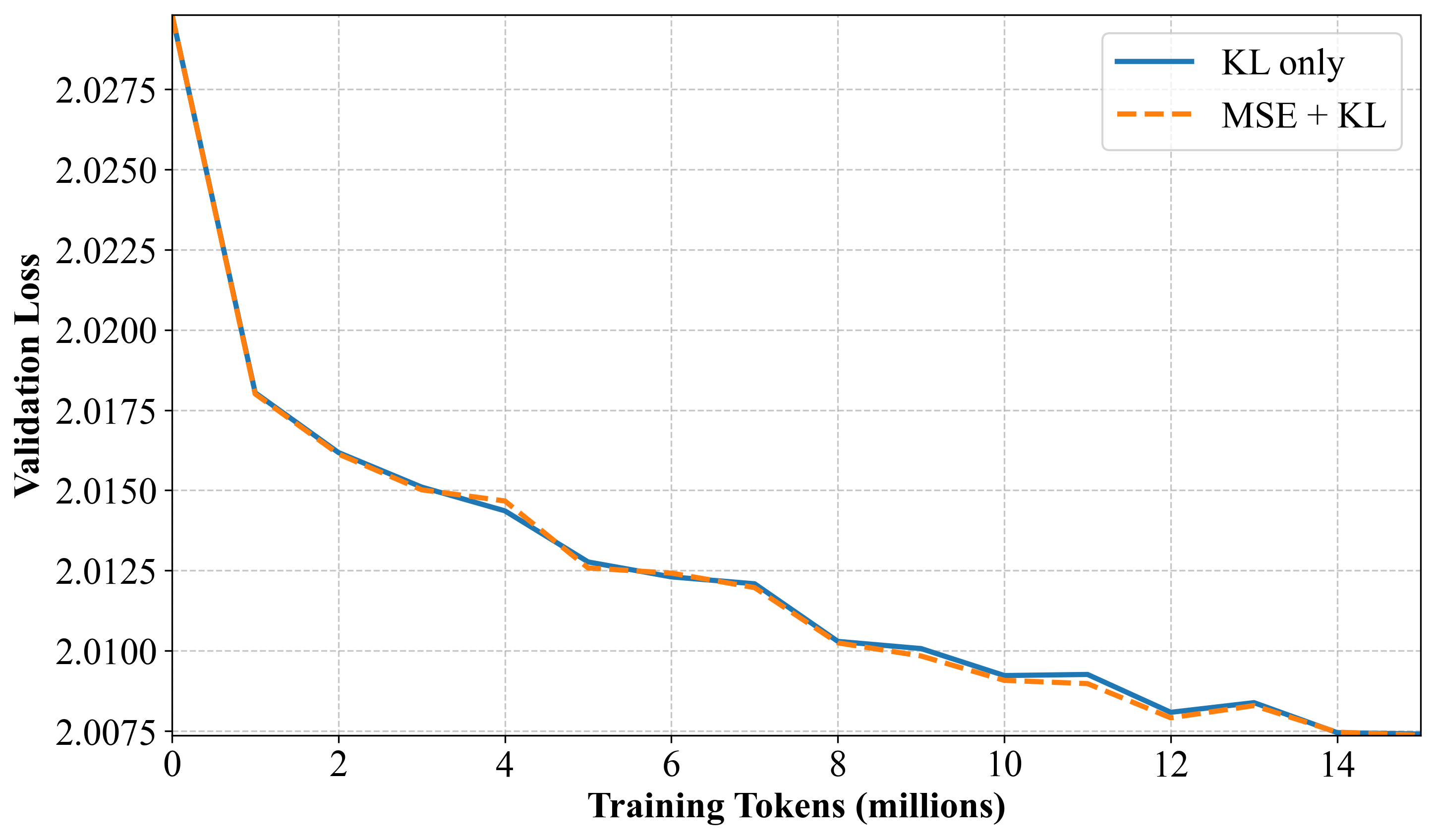}
        \caption{Validation loss comparison}
        \label{fig:kl_mse_vs_kl_b}
    \end{subfigure}
    \caption{Comparison of training with KL+MSE loss versus KL-only loss. There is virtually no difference in validation loss between the two methods, while KL only shows significantly worse MSE on the training set.}
    \label{fig:kl_mse_vs_kl}
\end{figure*}

\clearpage
\section{Alternative Adapter Experiments}
\label{app:alternative_adapter_experiments}

We explored additional mechanisms to further reduce cross-entropy loss beyond SAE fine-tuning. Specifically, we experimented with two lightweight adaptation methods applied after a pre-trained sparse autoencoder (SAE):

Low-Rank Linear Transform with Skip Connection: We introduced a low-rank linear layer defined as follows:
\[
y = x + UV x
\]
where $x$ is the SAE output, $U \in \mathbb{R}^{d\times r}$ and $V \in \mathbb{R}^{r\times d}$ are learnable matrices with rank $r \ll d$. Following LoRA-style initialization, $V$ is initialized to zero, ensuring the adaptation has no effect at the start of training.

Small MLP with Skip Connection: We tested a simple two-layer MLP with a small hidden dimension and ReLU activations, again with a residual connection:
\[
y = x + W_2\text{ReLU}(W_1 x + b_1) + b_2
\]
where $W_1 \in \mathbb{R}^{h\times d}$, $W_2 \in \mathbb{R}^{d\times h}$ (zero-initialized), $b_1 \in \mathbb{R}^h$, and $b_2 \in \mathbb{R}^d$ (zero-initialized), with hidden dimension $h \ll d$. The zero initialization of $W_2$ and $b_2$ ensures the residual connection initially passes through the input unchanged.

\begin{figure*}[h]
\centering
\begin{subfigure}[b]{0.48\textwidth}
\centering
\includegraphics[width=\textwidth]{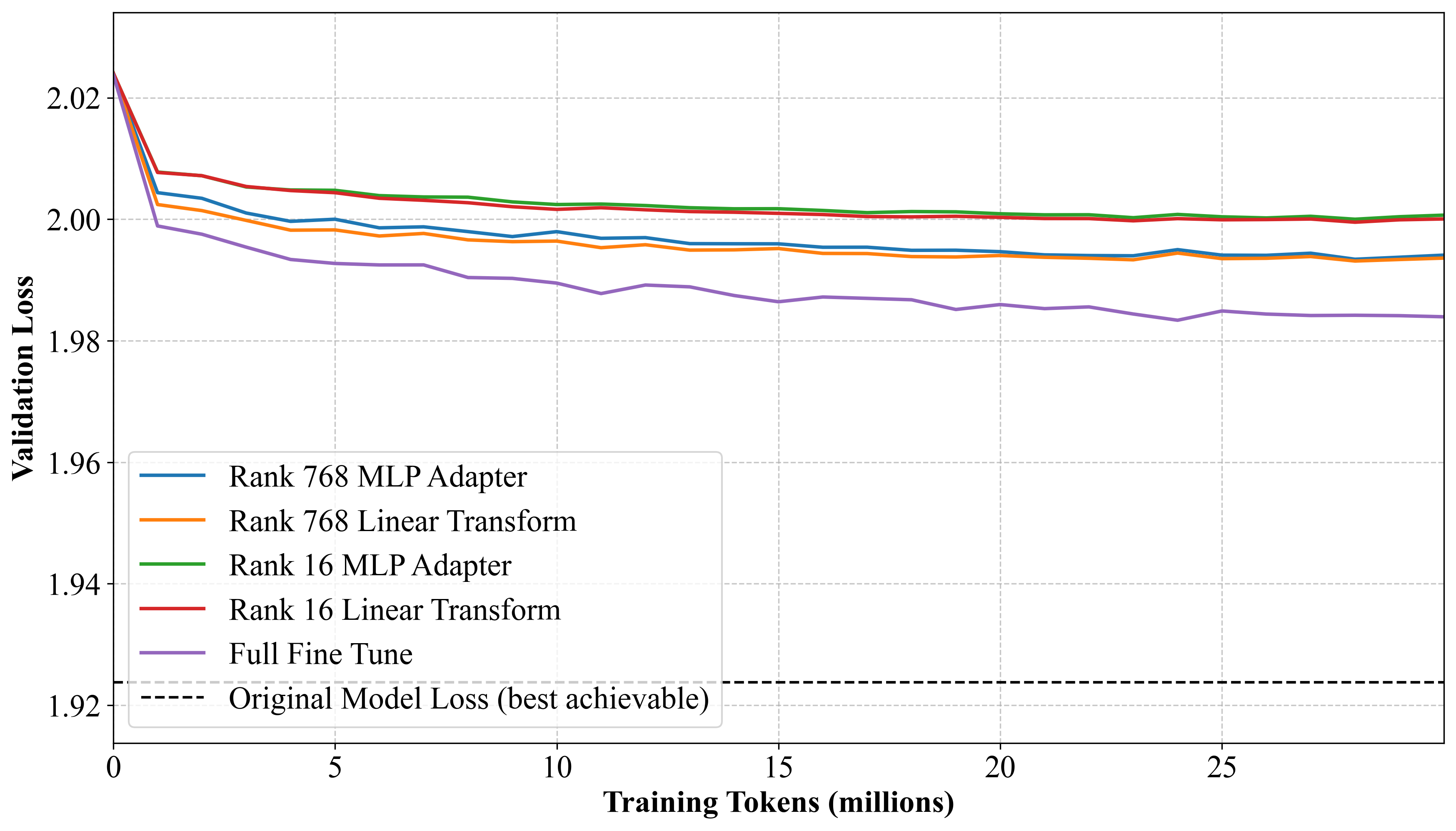}
\caption{Individual adaptation methods compared to SAE fine-tuning}
\label{fig:adaptation_methods_comparison}
\end{subfigure}
\hfill
\begin{subfigure}[b]{0.48\textwidth}
\centering
\includegraphics[width=\textwidth]{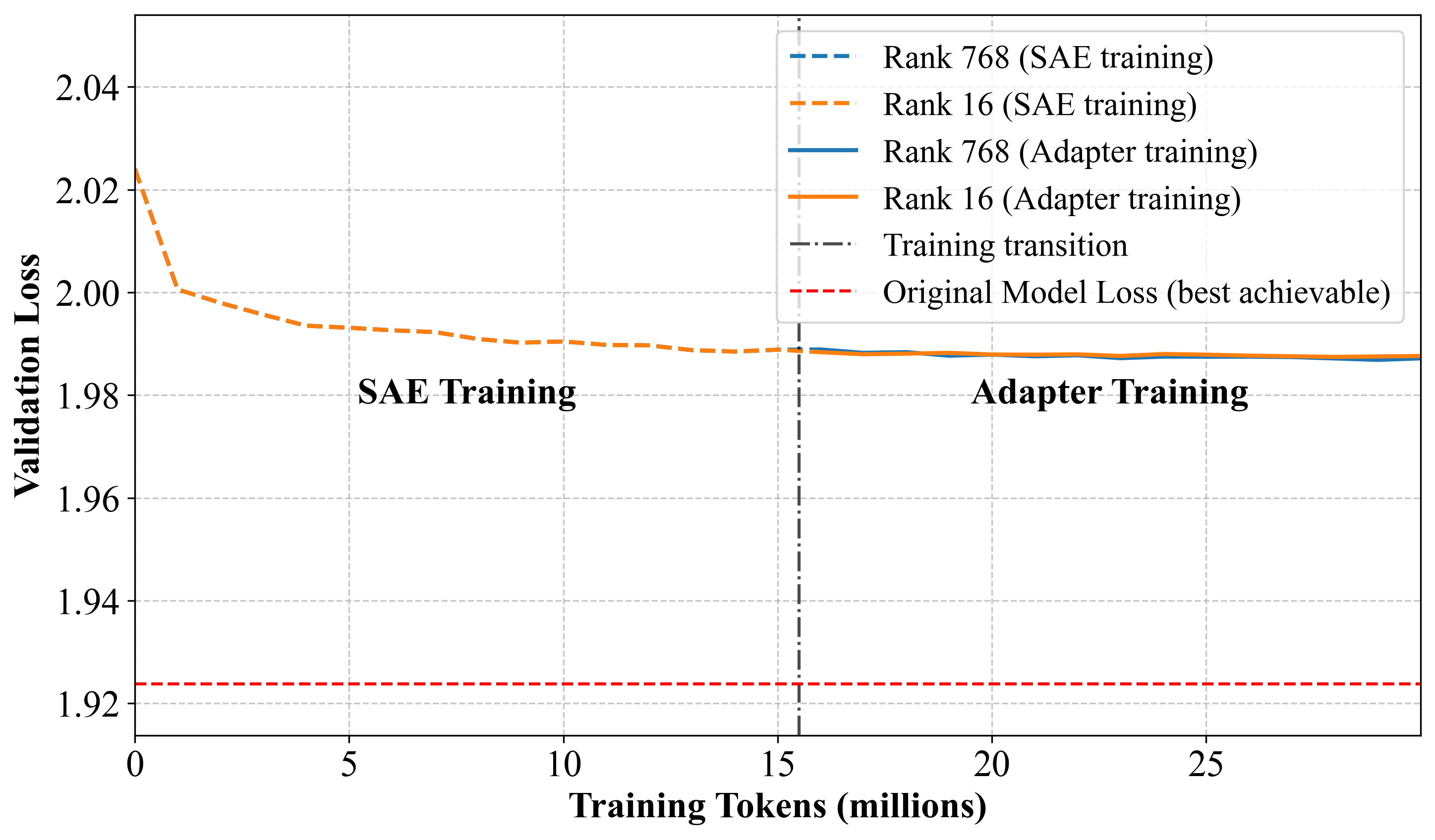}
\caption{Sequential two-stage training}
\label{fig:two_stage_training}
\end{subfigure}
\caption{Comparison of adaptation methods. SAE fine-tuning alone provides most improvements; additional methods yield minimal incremental gains.}
\label{fig:adapter_experiments}
\end{figure*}

Results: Both methods individually yielded significant improvements in cross-entropy loss—roughly 60\% of the improvement achievable via SAE fine-tuning alone (Figure~\ref{fig:adaptation_methods_comparison}).

To evaluate whether the gains from these methods would stack with SAE fine-tuning, we performed an additional two-stage fine-tuning experiment. First, we fine-tuned the SAE, and subsequently fine-tuned the low-rank linear transform or MLP on top of the already fine-tuned SAE. As shown in Figure~\ref{fig:two_stage_training}, we observed minimal additional improvement from this sequential approach, suggesting these methods primarily capture similar underlying improvements.

\textbf{Conclusion}: These results suggest that while lightweight adaptation methods can individually reduce cross-entropy loss, their benefits do not significantly stack with SAE fine-tuning. Therefore, to minimize complexity, we recommend using SAE fine-tuning alone in practice.

\clearpage
\section{LoRA Adapters on Sparse Autoencoders}
\label{app:sae_lora_experiment}

We additionally experimented with training LoRA adapters directly on the SAE weights, instead of performing a full fine-tune of the SAE parameters. Our results indicate that fully fine-tuning the SAE consistently yielded better performance. However, we found that even a very low-rank LoRA adapter (rank 2) could capture more than half of the performance gains achievable through a full fine-tune. This suggests that a substantial portion of the SAE reconstruction error is due to a relatively simple and easily correctable issue.

\begin{figure*}[!htb]
    \centering
    \includegraphics[width=0.85\linewidth]{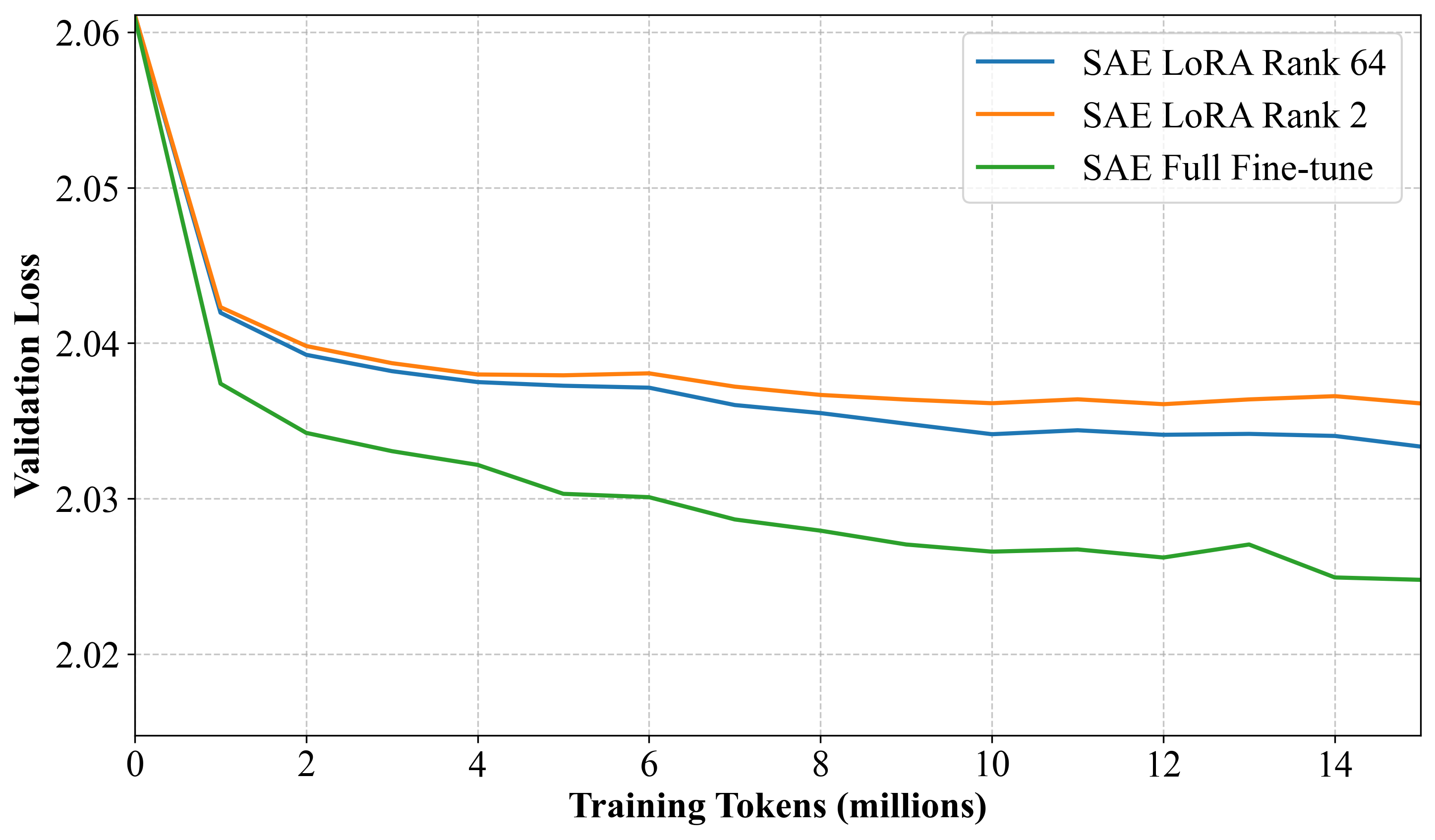}
    \caption{
        Comparison of a full fine-tune of the SAE versus rank 2 and rank 64 LoRA adapters.
    }
    \label{fig:sae_lora_experiment}
\end{figure*}

\section{ReLU Sparsity Penalty Adjustments}
\label{app:relu_sparsity_penalty_adjustments}

Using our adaptive balancing between KL and MSE losses, we would expect perfect transfer of sparsity penalties. However, we observed some deviations, possibly due to unshuffled activations. To ensure fair evaluation by maintaining consistent L0 sparsity levels across experiments, we implemented a simple dynamic sparsity penalty adjustment:

\begin{lstlisting}[language=Python]
def adjust_l1_penalty(current_l0, target_l0, l1_penalty):
    """Dynamically adjust L1 penalty to maintain target L0 sparsity."""
    adjustment_rate = 0.001
    if current_l0 < target_l0:
        l1_penalty *= (1 - adjustment_rate)
    else:
        l1_penalty *= (1 + adjustment_rate)
    return l1_penalty
\end{lstlisting}

This controller adjusts the L1 penalty during training to maintain the desired L0 sparsity level. While simple, we found this approach to be effective and stable across our experiments.

This sparsity control approach was adapted from an implementation shared by Glen Taggart.

\clearpage
\section{Batch Size Investigation}
\label{app:batch_size_investigation}

When training with KL divergence loss, activation shuffling is not possible, resulting in correlated activations from sequential tokens. To investigate the impact of batch size on this correlation, we conducted experiments comparing fine-tuning of 65K width TopK SAEs from the SAEBench suite using two different configurations: a batch size of 2 with sequence length 1024, and a batch size of 32 with the same sequence length. Both experiments were conducted over 25 million tokens. Analysis of SAEBench metrics revealed only minor, insignificant differences between the two batch sizes.

\begin{figure*}[!htb]
    \centering
    \includegraphics[width=0.85\linewidth]{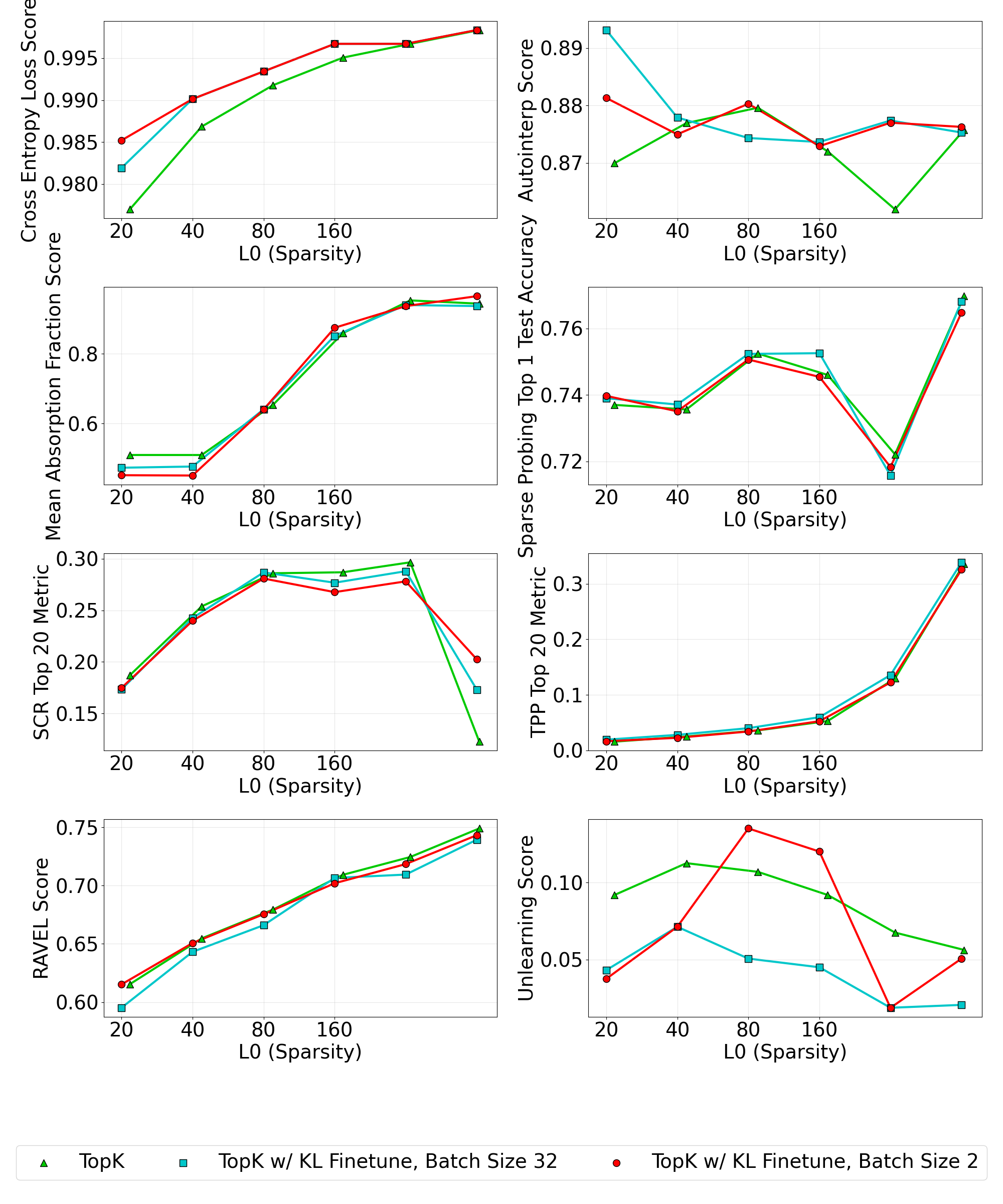}
    \caption{
    SAEBench results for 65K width TopK SAEs using KL+MSE finetuning with batch size 2 and 32.
    }
    \label{fig:batch_size_investigation}
\end{figure*}

\clearpage
\section{Further SAEBench Results}
\label{app:saebench_results}

\subsection{SCR and TPP, number of latents ablated hyperparameter sweep}
\label{app:saebench_results_scr_tpp_ablation}

The Spurious Correlation Removal (SCR) and Targeted Probe Perturbation (TPP) metrics each have a hyperparameter k that determines the number of SAE latents to ablate. In the main results, following the SAEBench paper, we present results using k=20. Here, we analyze the impact of this hyperparameter by sweeping across multiple k values and selecting the best performing configuration.

This analysis reveals more substantial improvements for fine-tuned TopK SAEs than previously shown, particularly at larger model widths. However, the general trends for ReLU SAEs remain consistent with our main findings: worse SCR and TPP scores. These results reinforce our conclusion that the benefits of KL+MSE fine-tuning vary significantly with architecture choice.

\begin{figure*}[h]
    \centering
    \begin{subfigure}[b]{0.48\textwidth}
    \centering
    \includegraphics[width=\textwidth]{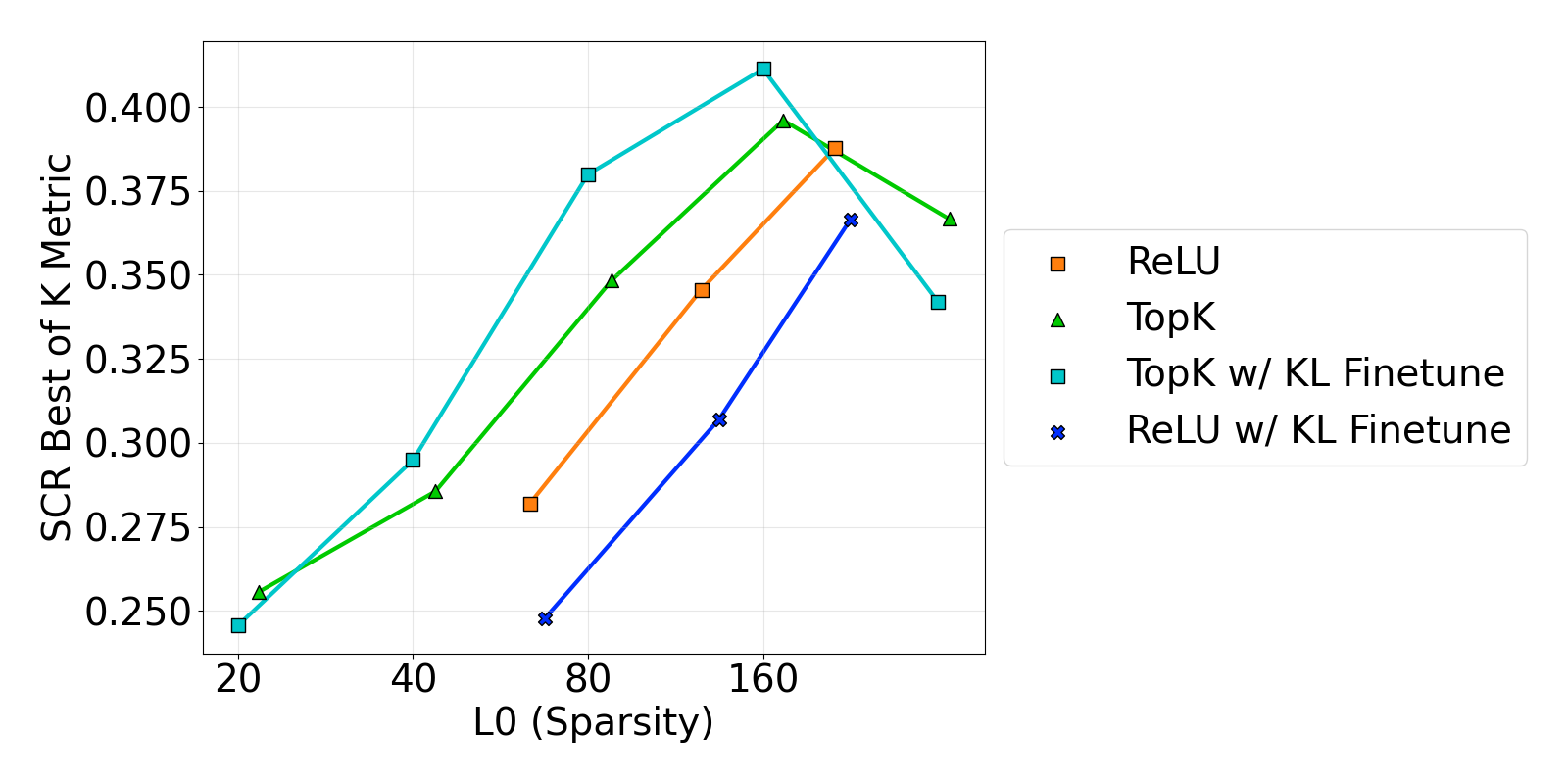}
    \caption{SCR, 65K width}
    \label{fig:scr_ablation_65k}
    \end{subfigure}
    \hfill
    \begin{subfigure}[b]{0.48\textwidth}
    \centering
    \includegraphics[width=\textwidth]{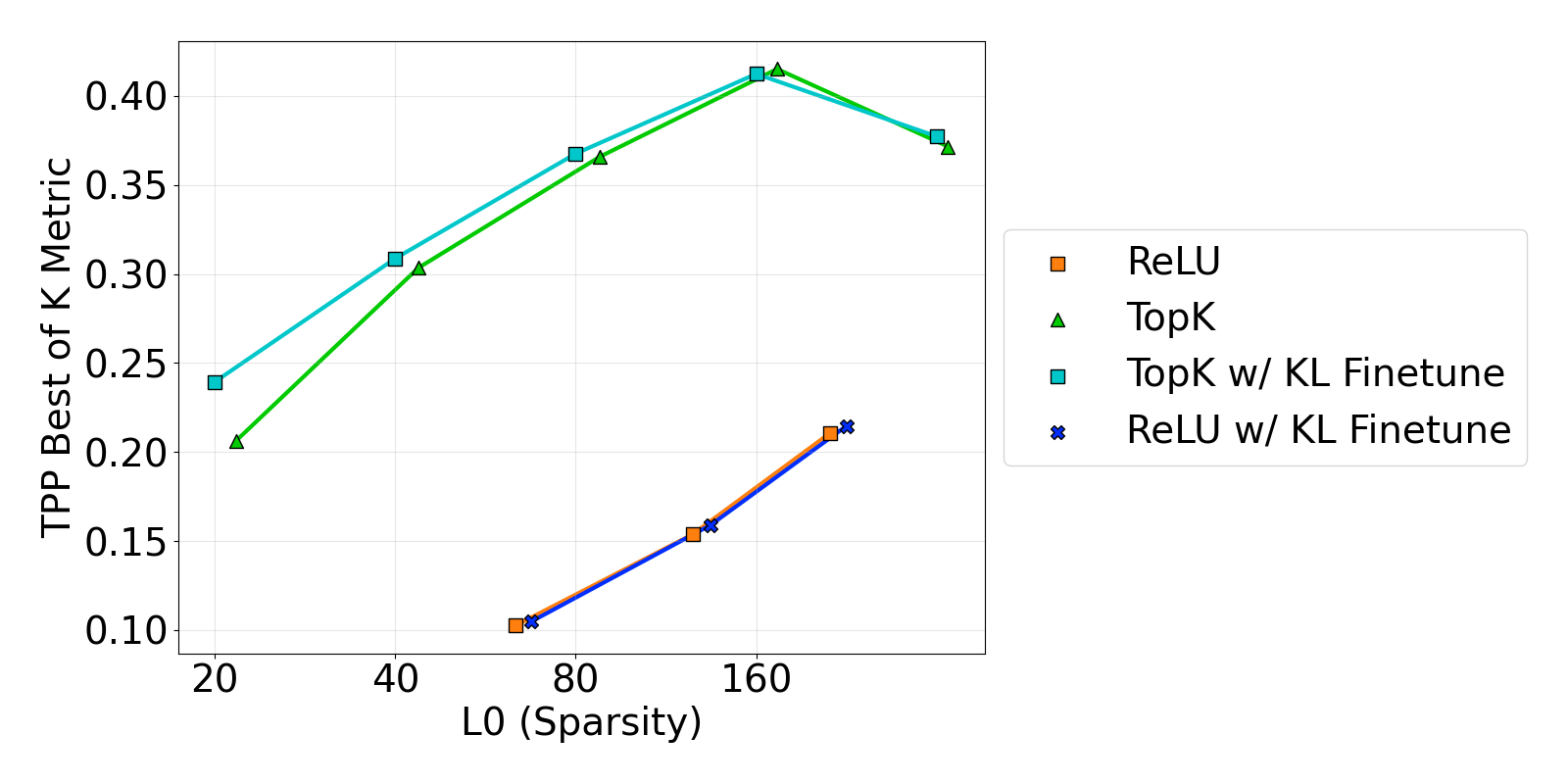}
    \caption{TPP, 65K width}
    \label{fig:tpp_ablation_65k}
    \end{subfigure}
    \caption{SCR and TPP, number of latents ablated hyperparameter sweep. We sweep across multiple k values and choose the best result obtained. We see more substantial improvements for fine-tuned TopK SAEs, especially larger sizes.}
    \label{fig:scr_tpp_ablation}
\end{figure*}

\begin{figure*}[h]
    \centering
    \begin{subfigure}[b]{0.48\textwidth}
    \centering
    \includegraphics[width=\textwidth]{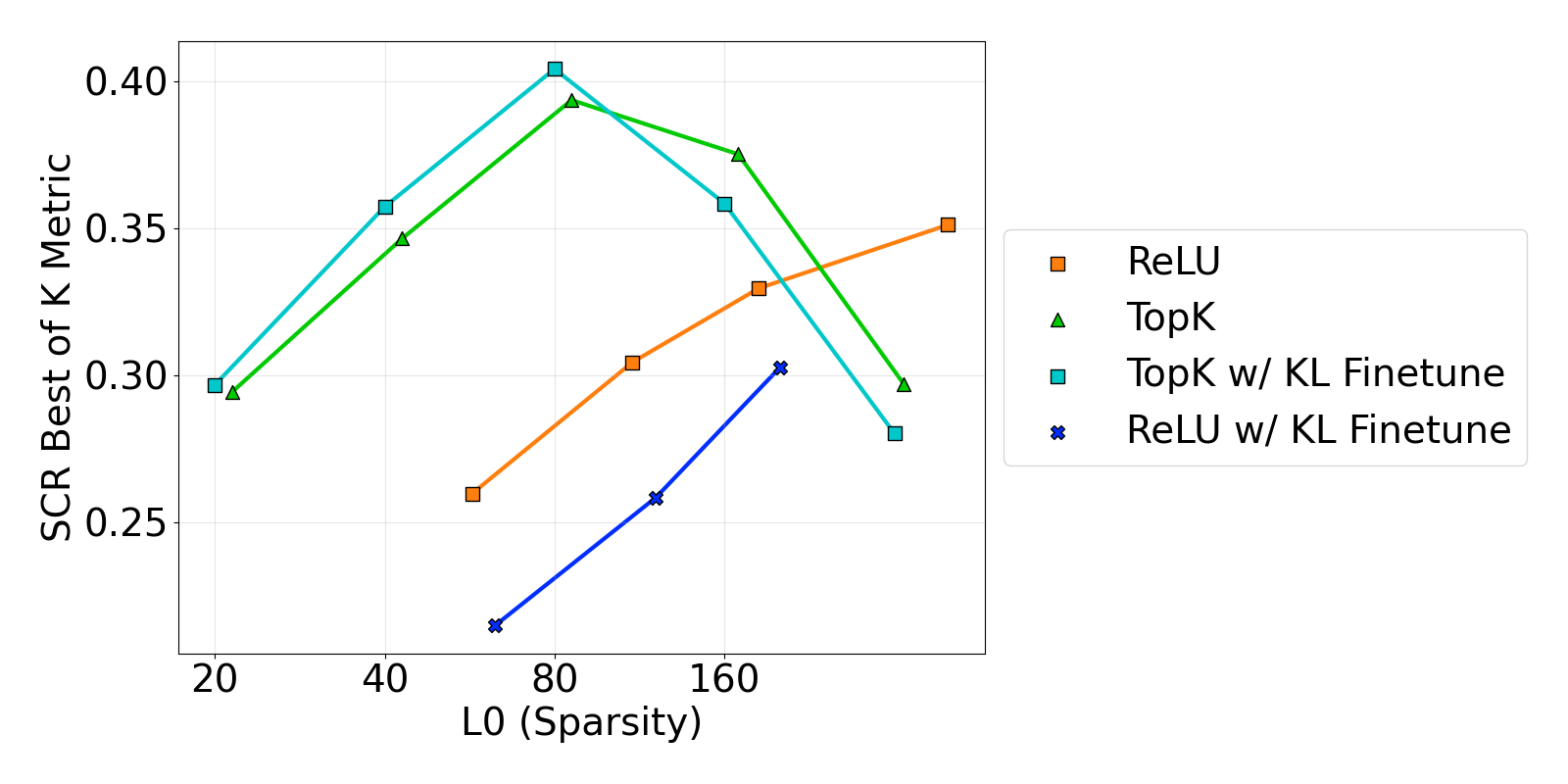}
    \caption{SCR, 16K width}
    \label{fig:scr_ablation_16k}
    \end{subfigure}
    \hfill
    \begin{subfigure}[b]{0.48\textwidth}
    \centering
    \includegraphics[width=\textwidth]{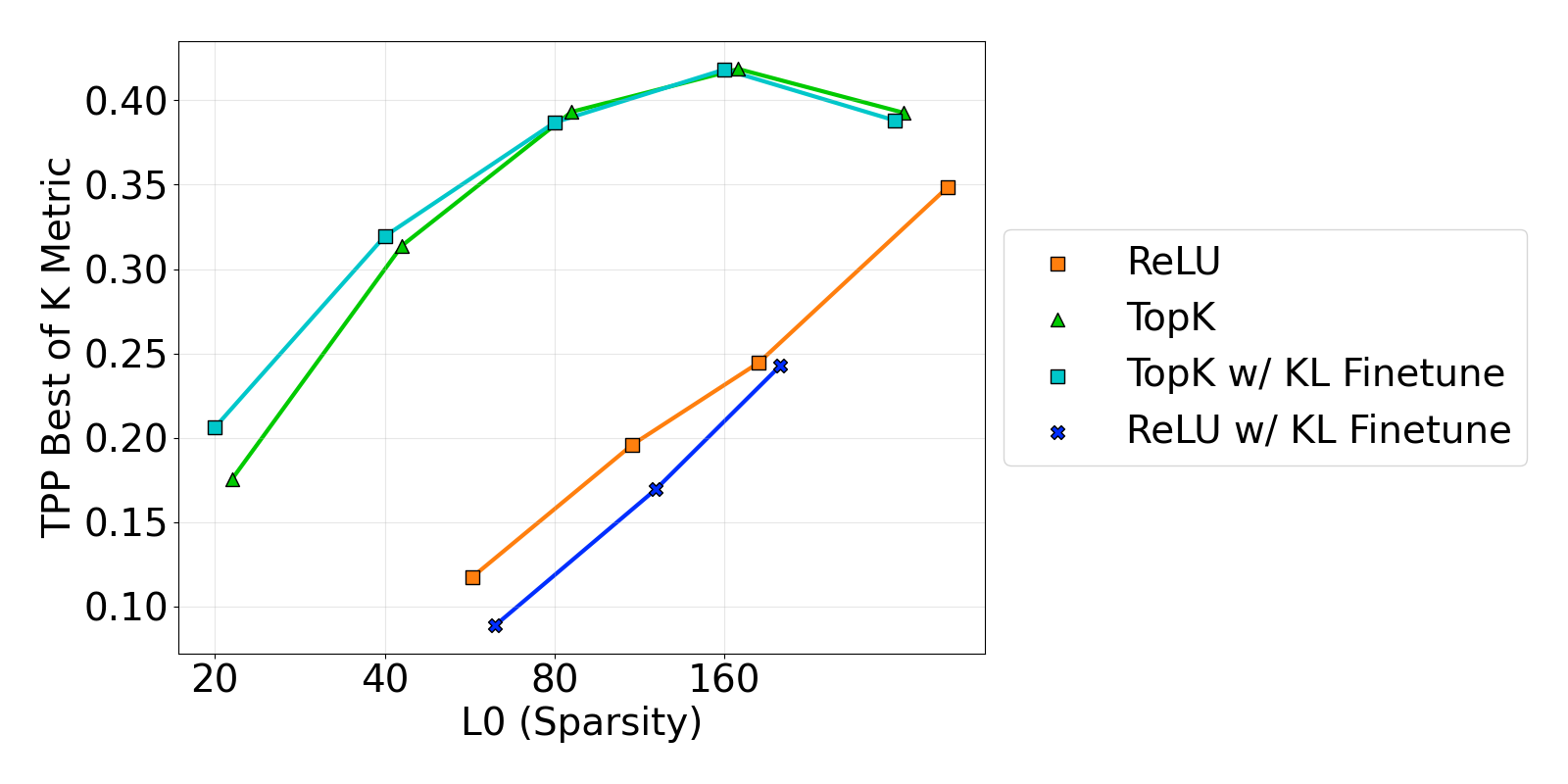}
    \caption{TPP, 16K width}
    \label{fig:tpp_ablation_16k}
    \end{subfigure}
    \caption{SCR and TPP, number of latents ablated hyperparameter sweep. We sweep across multiple k values and choose the best result obtained. We see more substantial improvements for fine-tuned TopK SAEs, especially larger sizes.}
    \label{fig:scr_tpp_ablation_16k}
\end{figure*}

\begin{figure*}[!htb]
    \centering
    \includegraphics[width=0.85\linewidth]{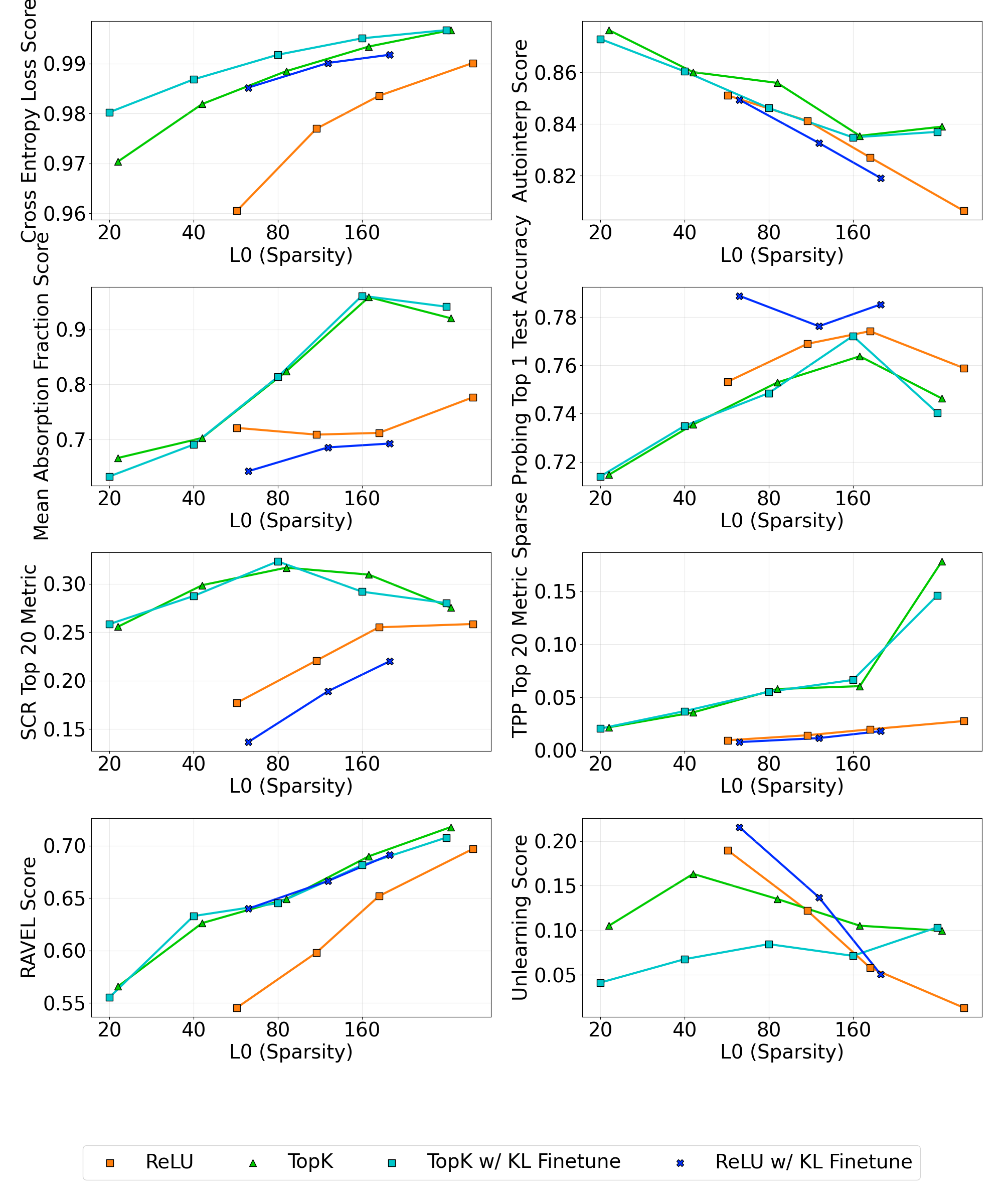}
    \caption{
    SAEBench results for KL+MSE finetuning (15M tokens) on 16k width SAEs on Gemma-2-2B.
    }
    \label{fig:saebench_16k}
\end{figure*}

\begin{figure*}
    \centering
    \includegraphics[width=0.85\linewidth]{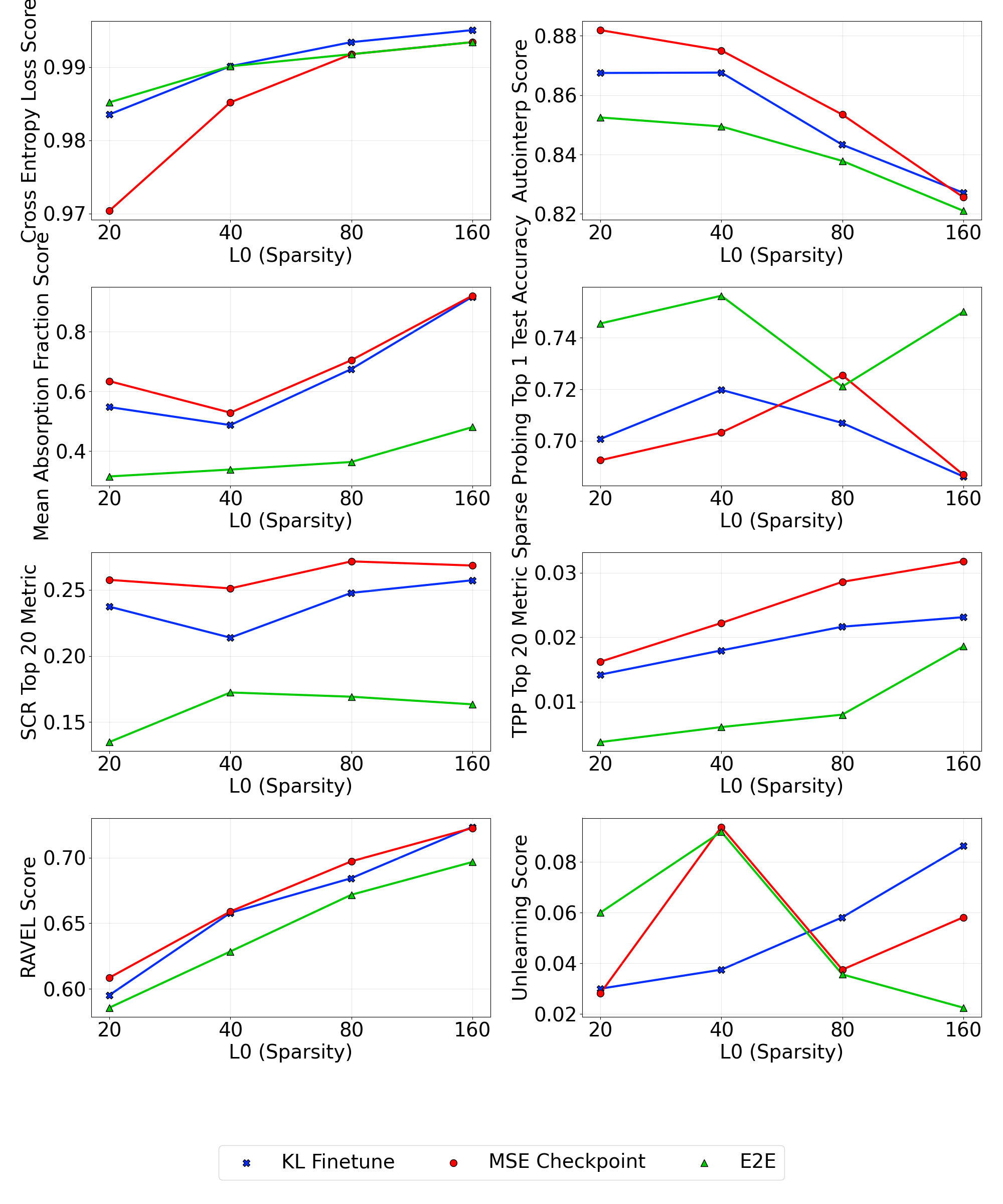}
    \caption{
    SAEBench results for SAEs trained using MSE only on 475M tokens, E2E training on 500M tokens, and MSE followed by KL+MSE finetuning on 25M tokens.
    }
    \label{fig:saebench_from_scratch}
\end{figure*}

\clearpage
\section{Weight Stability during KL+MSE fine-tuning}
\label{app:weight_stability}

We analyzed the stability of SAE weights during KL+MSE finetuning by measuring cosine similarities between pre- and post-finetuning weights for both encoder and decoder matrices. This analysis was performed across two architectures (TopK and ReLU) and two width configurations (16K and 65K).

Interestingly, we observed different patterns of weight stability across configurations. In both TopK and 65K ReLU models, the decoder weights showed greater changes (lower cosine similarities) compared to encoder weights during finetuning. However, the 16K ReLU model exhibited an unexpected pattern where encoder weights underwent more substantial changes than decoder weights. This difference in behavior between the 16K and 65K ReLU models suggests that model width may play a significant role in how weights adapt during KL+MSE finetuning.

The plots below show the 25th-75th percentile ranges of cosine similarities for the encoder and decoder weights over the course of finetuning. Higher values indicate greater stability, with a cosine similarity of 1.0 representing perfectly preserved weights.

\begin{figure*}[h]
    \centering
    \begin{subfigure}[b]{0.48\textwidth}
    \centering
    \includegraphics[width=\textwidth]{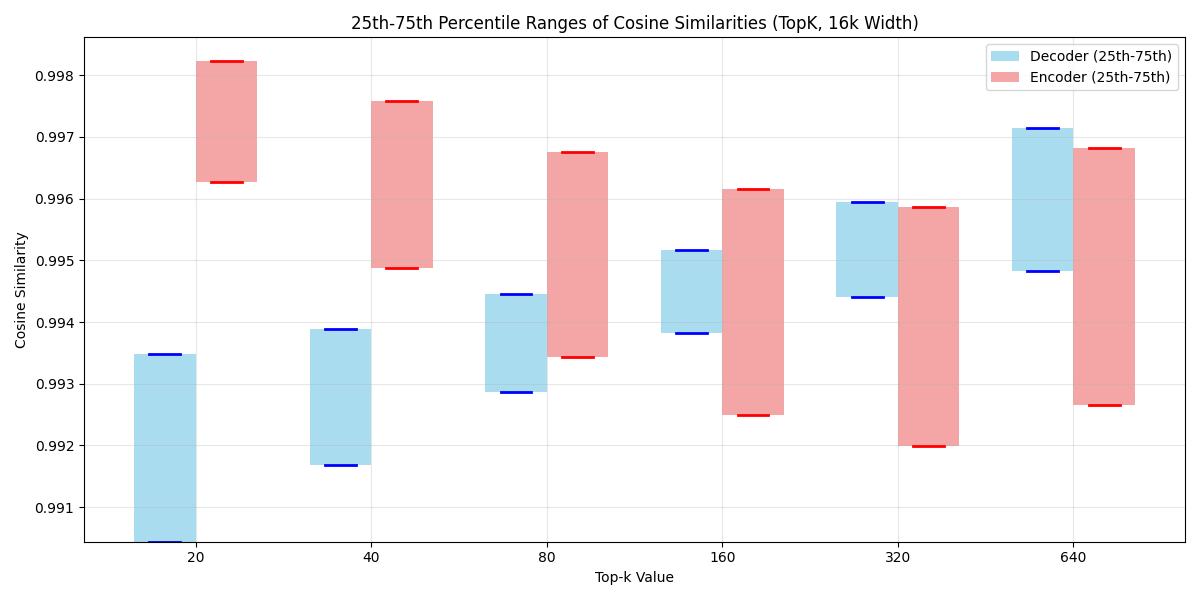}
    \caption{16K width TopK SAEs}
    \label{fig:topk_cos_sim_analysis_width_2pow14}
    \end{subfigure}
    \hfill
    \begin{subfigure}[b]{0.48\textwidth}
    \centering
    \includegraphics[width=\textwidth]{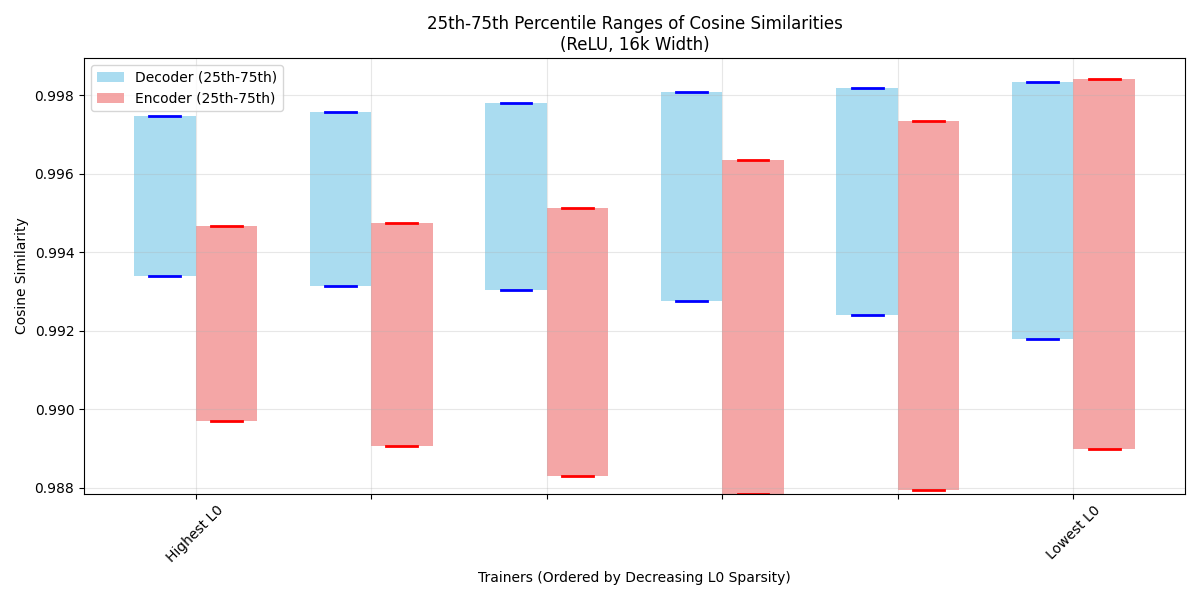}
    \caption{16K width ReLU SAEs}
    \label{fig:relu_cos_sim_analysis_width_2pow14}
    \end{subfigure}
    \caption{Weight stability during KL+MSE finetuning. The plots show the 25th-75th percentile ranges of cosine similarities for the encoder and decoder weights of the TopK and ReLU SAEs. The 16K ReLU model exhibits an unexpected pattern where encoder weights underwent more substantial changes than decoder weights.}
    \label{fig:weight_stability}
\end{figure*}

\begin{figure*}[h]
    \centering
    \begin{subfigure}[b]{0.48\textwidth}
    \centering
    \includegraphics[width=\textwidth]{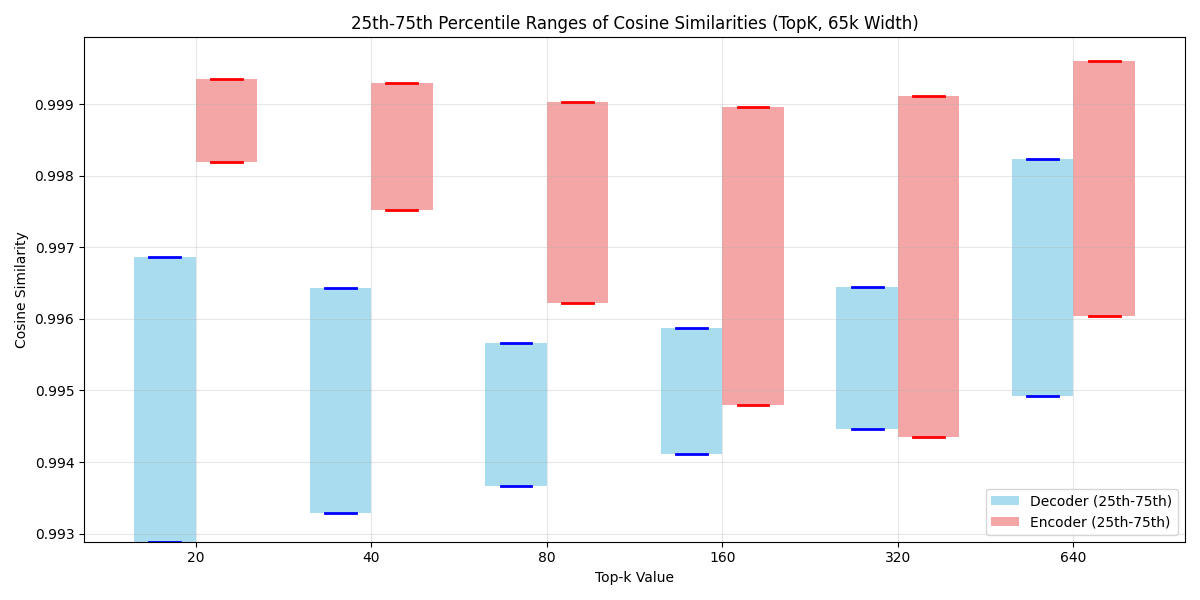}
    \caption{65K width TopK SAEs}
    \label{fig:topk_cos_sim_analysis_width_2pow16}
    \end{subfigure}
    \hfill
    \begin{subfigure}[b]{0.48\textwidth}
    \centering
    \includegraphics[width=\textwidth]{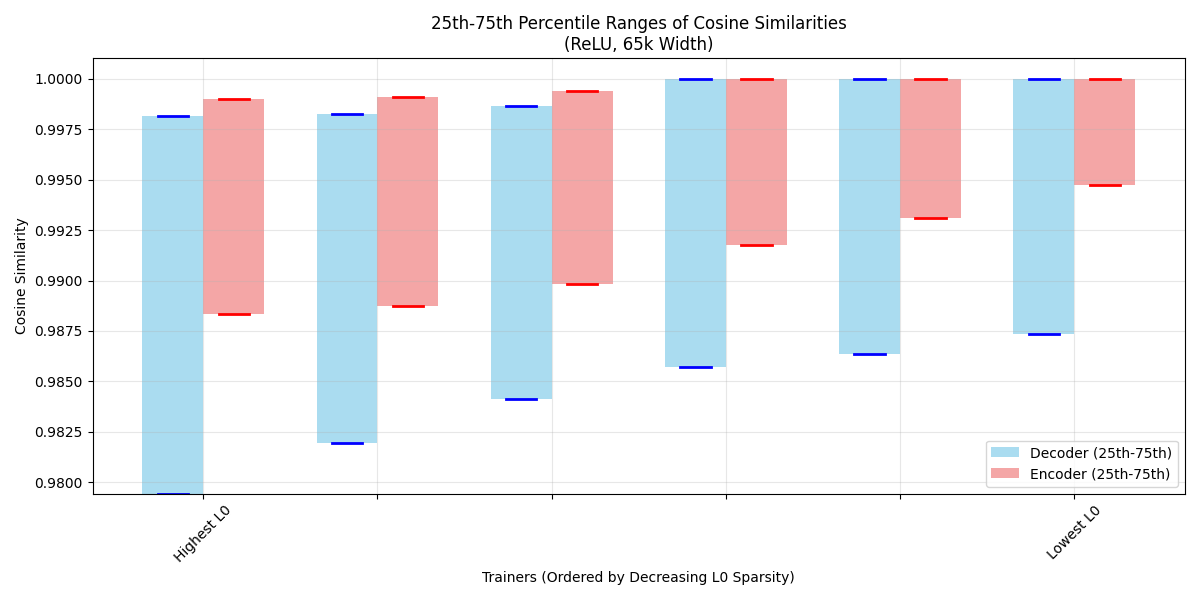}
    \caption{65K width ReLU SAEs}
    \label{fig:relu_cos_sim_analysis_width_2pow16}
    \end{subfigure}
    \caption{Weight stability during KL+MSE finetuning. The plots show the 25th-75th percentile ranges of cosine similarities for the encoder and decoder weights of the TopK and ReLU SAEs.}
    \label{fig:weight_stability_65k}
\end{figure*}

\end{document}